\documentclass[hidelinks]{article}
\usepackage{arxiv}
\usepackage{algorithm}
\usepackage{algpseudocode}
\usepackage{algorithmicx}
\usepackage{enumerate}
\usepackage{tikz}
\usepackage{amsmath}
\usepackage{booktabs,tabularx}

\usepackage{pgfplots}
\usepackage{subcaption}
\usepackage{soul}
\usepgfplotslibrary{groupplots}
\usepackage{amsmath,amssymb,amsfonts}

\begin{document}
\title{Adversarial Attacks on Locally Private Graph Neural Networks}
\author{%
  Matta Varun\\
  Indian Institute of Technology Kharagpur, India\\
  \texttt{varunkaleb@gmail.com}   \And
  Ajay Kumar Dhakar\\
  Indian Institute of Technology Kharagpur, India\\
  \texttt{ajaydhaker2002@gmail.com}   \And
  Yuan Hong \\
  University of Connecticut, USA\\
  \texttt{yuan.hong@uconn.edu}   \And
  Shamik Sural \\
  Indian Institute of Technology Kharagpur, India\\
  \texttt{shamik@cse.iitkgp.ac.in} 
}
\date{}
\renewcommand{\headeright}{}
\renewcommand{\undertitle}{}
\maketitle

\begin{abstract}
Graph neural network (GNN) is a powerful tool for analyzing graph-structured data. However, their vulnerability to adversarial attacks raises serious concerns, especially when dealing with sensitive information. Local Differential Privacy (LDP) offers a privacy-preserving framework for training GNNs, but its impact on adversarial robustness remains underexplored. This paper investigates adversarial attacks on LDP-protected GNNs. We explore how the privacy guarantees of LDP can be leveraged or hindered by adversarial perturbations. The effectiveness of existing attack methods on LDP-protected GNNs are analyzed and potential challenges in crafting adversarial examples under LDP constraints are discussed. Additionally, we suggest directions for defending LDP-protected GNNs against adversarial attacks. This work investiagtes the interplay between privacy and security in graph learning, highlighting the need for robust and privacy-preserving GNN architectures.
\end{abstract}

% \ccsdesc[300]{Security and privacy~Privacy-preserving protocols}
% \ccsdesc[300]{Computing methodologies~Machine learning algorithms}


\section{Introduction}
\label{sec:intro}

Graph Neural Networks (GNNs) \cite{gnnPaper} have revolutionized the field of machine learning by enabling effective learning from graph-structured data. Their ability to capture complex relationships among the nodes of a graph has led to significant advancements in tasks like node classification, community detection, and link prediction \cite{kipf_gnn_gcn}, \cite{veličković2018_gans}. However, widespread adoption of GNNs in domains handling sensitive information, such as social networks or financial transactions, necessitates robust security measures.

Adversarial attacks pose a significant threat to the security of GNNs. Malicious actors can manipulate the input data inducing the model to make incorrect predictions, potentially leading to disastrous consequences. Recent research has extensively investigated adversarial attacks on basic GNNs, exploring various methods to perform adversarial perturbations that can alter the model's output \cite{adversarials_on_gnn}, \cite{graph_injection_attack}. These attacks highlight the vulnerability of GNNs to carefully crafted manipulations, raising concerns about their reliability in real-world applications. The attacker can also carry out attacks such as model inversion \cite{modelInversaionAgainstGNN} or membership-inference attacks, which would violate privacy of the trained graph data. In addition, the possibility of server being compromised cannot be excluded, as a consequence of which private and sensitive training graph data can be leaked.

Based on their objectives, adversarial attacks on GNNs can be broadly classified into two primary categories. The first category encompasses attacks that aim to compromise the model's accuracy and overall performance on graph-related tasks. These attacks manipulate the input graph in a way that the GNN model is compelled to make incorrect predictions, such as mis-classifying a node or predicting wrong labels for an entire graph. The second category focuses on attacks that seek to leak private information embedded within the graph data or to undermine the security of the graph information being processed by the GNN. These attacks aim to compromise the confidentiality and integrity of the graph data itself, potentially revealing sensitive relationships, attributes, or even the composition of the training dataset. In the realm of attacks targeting model accuracy, further distinctions can be made based on the stage at which the attack occurs, namely, during the training phase (poisoning attacks) or during the inference phase (evasion attacks), as well as the attacker's specific goal. It could either be to misclassify a particular target instance (targeted attacks) or to degrade the overall performance of the model across a range of instances (untargeted attacks). Understanding this fundamental categorization is essential for navigating the complexities of adversarial threats against GNNs and for developing appropriate countermeasures.

The scenario becomes intricate when we consider Locally Differentially Private (LDP) GNNs as proposed in \cite{lpgnn}. LDP algorithms offer strong privacy guarantees by injecting noise into the data during training, thereby ensuring that the model output does not reveal any specific detail about individual data points, thus protecting user privacy \cite{Algorithmic_foundations_of_DP}. While LDP offers significant privacy benefits, the inherent noise introduced into the data could possibly create new vulnerabilities to adversarial manipulation.
% Hence the study of adversarial attacks on GNNs has gained significant importance due to increasing deployment of these models in real-world applications, many of which are safety-critical and involve sensitive data. 
It raises serious concerns about their reliability in domains such as financial fraud detection, where malicious actors could manipulate transaction graphs to evade detection. 
% Similarly, in risk management systems that rely on GNNs to assess the risk associated with individuals or entities within a network, adversarial perturbations could lead to flawed assessments with potentially severe consequences. 
% The ability of attackers to spread disinformation or manipulate public opinion on social media platforms by subtly altering the connections and features within social network graphs processed by GNNs also underscores the urgency of addressing these vulnerabilities. 
Beyond immediate security concerns, adversarial attacks also serve as a valuable tool for probing the fundamental properties and limitations of GNN models, helping to understand their decision boundaries and the factors influencing predictions. Ultimately, a deeper understanding of these vulnerabilities is crucial for the development of more robust GNN models and the design of effective defense mechanisms that can safeguard their performance along with privacy of the data they process. 

In this paper, we explore how attackers can exploit the privacy-preserving mechanisms of LDP GNNs (hereinafter referred to as LPGNN) to craft adversarial perturbations. These perturbations can remain undetected by the privacy noise while significantly altering the model's predictions, potentially compromising the model's accuracy and overall functionality. Specifically, we consider four types of attack, namely, \textit{Node Injection Attack}, \textit{Label-Flipping Attack}, \textit{Inference Attack} and \textit{Poisoning Attack}.  By investigating this critical research gap, we aim to contribute to a deeper understanding of the security landscape surrounding LDP GNNs. Our work not only identifies potential vulnerabilities but also provides a direction towards mitigation strategies for building more robust privacy-preserving graph learning models. This will ultimately lead to the development of secure and reliable GNNs suitable for real-world applications with stringent privacy requirements.

% The rest of the paper is organized as follows. In Section \ref{sec:prelims}, we introduce some of the background material while the main attacks proposed by us form the content of Section \ref{sec:proposedattacks}. Experiments are conducted in Section \ref{sec:experiments}. Related work is reviewed in Section \ref{sec:related} and we finally conclude in Section \ref{sec:concl}, providing directions for future work in this domain.

\section{Preliminaries}
\label{sec:prelims}
Here, we provide an introduction to Graph Neural Networks, Local Differential Privacy, as well as LPGNN, establishing the necessary foundation for understanding adversarial attacks on LDP GNNs.

% \subsection{Graph Neural Networks}
% \label{subsec:GNN}

GNNs constitute a useful class of neural networks designed to operate on graph-structured data \cite{gnnPaper}\cite{gnn_Explain}. Unlike traditional neural networks that process sequential data, GNNs can effectively embed the relationships and dependencies among the nodes of a graph. 
% Here's a formal definition:
% A graph $G$ can be represented as $G = (V, E)$, where $V$ is a set of nodes representing entities within the data and $E$ is a set of edges capturing the connections or relationships between these nodes. Each node $v \in V$ can have associated features denoted by $x_v$.
A message passing scheme is employed in GNNs to iteratively aggregate information from a node's neighborhood. In every iteration, a message function computes a new representation for each node based on its own features and the features of its neighbors. These messages are then aggregated using an aggregation function and combined with the node's current representation to update its state. The process is repeated for multiple steps, allowing the GNN to learn informative node representations that encode not only the node's intrinsic features but also the structural information of the graph.
By stacking multiple GNN layers, the network can progressively learn increasingly complex representations of the nodes, capturing the intricate relationships within the graph structure. These learned representations can then be used for various tasks like node classification, link prediction, and community detection.
% A mathematical formulation of a typical GNN layer is given below.

% \textbf{Message Passing}:
% \begin{equation*}
%     M_v^{(t)} = MESSAGE(h_v^{(t-1)}, \{h_u^{(t-1)} : u \in N(v)\}) \forall v \in V
%     % \label{eq:messagepassing}
% \end{equation*}
% where $M_v^{(t)}$ is the message received by node $v$ in layer $t$, $h_v^{(t-1)}$ is the hidden representation of node $v$ in layer $t-1$, $N(v)$ denotes the set of neighboring nodes of $v$, $MESSAGE$ defines how the message is computed.

% \textbf{Aggregation}:
% \begin{equation*}
%     h^t_V = AGGREGATE_t( \{ M_v^{(t)} : v \in V\})
%     % \label{eq:aggregation}
% \end{equation*}
% This function aggregates the messages received by each node.

% \textbf{Update}:
% \begin{equation*}
%     h_v^{(t)} = UPDATE_t(h_v^{(t-1)}, h^t_V( \{ M_v^{(t)} : v \in V\}))
%     % \label{eq:update}
% \end{equation*}
% where, $h_v^{(t)}$ is the updated hidden representation of node $v$ in layer $t$ and $UPDATE_t$ defines how the node's previous state is combined with the aggregated messages in layer $t$. $UPDATE_t(.)$ is a trainable non-linear function for layer $t$. At the start, $h^0_v = x_v$, i.e., the initial embedding of $v$ is its feature vector $x_v$, and the last layer generates a $c$-dimensional output followed by a softmax layer to predict node labels in a $c$-class node classification
% task.

% \subsection{Local Differential Privacy}
% \label{subsec:LDP}

Local Differential Privacy \cite{kasiviswanathan2011can, rappor, ieee_tp} is a formal approach for ensuring privacy guarantees when collecting and analyzing sensitive data. It differs from traditional Differential Privacy (DP) \cite{dwork2006calibrating, Algorithmic_foundations_of_DP, Privacy_via_dist_noise_gen}, where data perturbation happens on the server, which poses a risk as the server can be compromised and may not be trusted by the clients in certain situations. LDP offers a stronger guarantee than traditional DP by perturbing the data at individual data holder devices before being aggregated. This way, the final analysis output preserves valuable statistical insights without revealing any private information about specific data points.

Data holders use a special algorithm to inject carefully calibrated noise into their data. This noise is mathematically guaranteed to satisfy the core principle of LDP: $\epsilon$-differential privacy. An LDP mechanism is considered $\epsilon$-differentially private if the output distribution barely changes when any single data point is added or removed from the dataset. 
% Formally, for any neighboring datasets (differing by one record) $D$ and $D'$, and for any output $O$, the probability of the mechanism outputting $O$ on $D$ is at most $e^\epsilon$ times the probability of outputting $O$ on $D'$, i.e.,
% \begin{equation}
% Pr[M(D) = O] \leq e^{\varepsilon} \cdot Pr[M(D') = O]
% \end{equation}
% where $\epsilon$ is also known as the privacy budget. 
The parameter $\epsilon$ controls the privacy-utility trade-off. Lower $\epsilon$ values offer stronger privacy guarantees by adding more noise, but this can also obscure the underlying patterns in the data.
The usefulness of LDP lies in its ability to protect individual privacy while enabling accurate statistical analysis. By injecting noise at the source, LDP prevents attackers from inferring private details about any single data point, even if they have access to the aggregated output.

% \subsection{Locally Private Graph Neural Networks}
% \label{subsec:LPGNN}
% \textbf{This section will contain the concepts introduced by the work \cite{lpgnn}. This model of GNN is herewith referred to as LPGNN in the rest of the paper.}

Recently, Sajadmanesh and Gatica-Pere~\cite{lpgnn} proposed a locally private graph neural network model, called LPGNN. The aim of LPGNN is to train Graph Neural Networks while preserving the privacy of node features and labels. The authors devise a framework combining multiple techniques to enable effective learning with formal privacy guarantees under LDP. LPGNN integrates various privacy-preserving components such as the Multi-Bit Mechanism for private feature collection, KProp for denoising, Randomized Response for label perturbation, and DROP (Label Denoising with Propagation) for robust training. 
% The following subsections describe each of these components in brief.
A Multi-Bit Encoder is executed on the user side, which perturbs the private feature vector and encodes it into a compact binary vector that can be transmitted to the server. The encoder ensures that the private node feature vector \( \mathbf{x} \) is converted into a vector \( \mathbf{x}^* \) which leaks minimal information while satisfying \( \epsilon \)-LDP. The encoding step requires only a single pass over the user's data, offering efficiency and reduced communication overhead. Once the encoded vector \( \mathbf{x}^* \) reaches the server, the Multi-Bit Rectifier transforms it into an unbiased perturbed feature vector \( \mathbf{x}' \). This process removes the statistical bias introduced by encoding, while still maintaining the privacy guarantees of LDP.

The authors also introduce a KProp Layer that performs simple feature aggregation over a K-hop neighborhood to denoise the input feature vectors before feeding them into the GNN. To  protect node label privacy, LPGNN applies a randomized response technique. With a probability \( p \), each node label \( y \) is flipped to a randomly selected label \( y' \). 
Such a perturbation ensures label privacy while still allowing the model to learn from noisy labels.
To counter the effects of label noise, the authors propose DROP, which estimates the true label distribution, avoiding overfitting and maintaining predictive performance without access to clean labels.

\section{Proposed Attacks}
\label{sec:proposedattacks}
We now introduce the different forms of adversarial attacks we have designed and conducted against LPGNNs. First, we discuss the prospective aims of an adversary while attacking the GNN. An adversary may want to compromise the utility and performance of a GNN trained on the current graph dataset. 
This can manifest in the form of reduced model accuracy, leading to sub-par performance. The adversary would be able to perform this by being an active part of the entire GNN training process, or they can perform it indirectly via methods such as poisoning of the dataset nodes. 
It is also possible for the adversary to want to break through the privacy guarantees offered by the LPGNNs, thereby compromising the protection offered to the sensitive graph node data. This can be carried out in the form of poisoning attacks, inference attacks, or a hybrid form of attack. Broadly speaking, these are the two main aims an adversary can possess while attacking an LPGNN.

We initially discuss two classical attacks, namely, Node-injection and Label-flipping, which are intended to degrade the performance of a GNN. We analyze the further impact these attacks have on LPGNN, as it is already known that LPGNN provides a comparatively poorer performance than a GNN trained on the same graph dataset. 
Next, we carry out an inference attack based on some assumption about the graph data. 
We study the outcomes of this attack across various settings and justify our findings. 
Finally, we carry out another poisoning attack, this time aimed at breaking the privacy guarantee of the LDP mechanism in LPGNN.

\subsection{Classical Attacks}
\label{subsec:classical_attacks}

We begin with two existing forms of attacks against GNNs, which have been researched and tested against typical non-private (i.e., non-LDP) graph neural networks. The impact of these attacks on traditional GNNs is well-known and already established. However, their effect in the locally private version of GNNs is not yet studied and analyzed. Hence, we design and conduct the following types of classical attacks against LPGNNs.

\subsubsection{\textbf{Node Injection Attack}}
\label{subsubsec:node_injection_attack}
% \hfill \vspace{0.5mm} \break
Injecting nodes with malicious intent into a graph structured data is a well-known form of attack \cite{effectiveNodeInjectionAttacks, singleNodeInjectionAttacks, advAttackOnGNNviaNodeInjections, GANI_NodeInjections, classificationOptNodeInjectionAttack}. The adversary crafts a node and inserts it into the graph data at a location with strategically formed edges with neighboring nodes. In the non-private setting of GNNs, these attacks are aimed at impacting the resultant performance of the GNN and increasing the node mis-classification rate. 
% The adversary crafts a node and inserts it into the graph data at a strategic location with strategically formed edges with neighboring nodes. In the non-private setting of GNNs, these attacks are aimed at impacting the resultant performance of the GNN and increasing the node mis-classification rate.
The work of  Zugner et al. \cite{Zugner_2018} shows how injecting sufficient amount of nodes at strategic locations can completely undermine the performance of a trained GNN with the mis-classification rate increasing drastically as a result of these attacks. 
% We know that introducing a locally differentially private mechanism into GNNs already comes at an accuracy cost. 
% We conducted node injection attacks against LPGNNs, measuring the classification rates after each attack. 
Below we describe how the attack is conducted on LPGNN.

\begin{figure*}[t]
    \centering
    \includegraphics[width=1\linewidth]{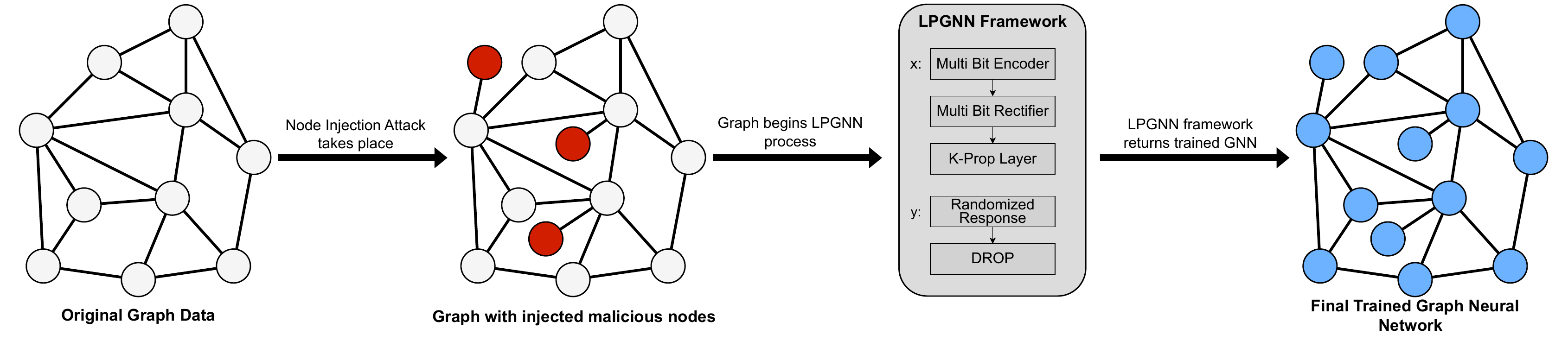}
    \captionsetup{justification=centering}
    % \caption{Visualization of Node Injection Attack on LPGNN: There are $n$ nodes in the graph, each of which holds sensitive data. The attack targets nodes with high degrees, injecting a node with a direct edge to the targeted nodes. This aims to mislead and spread random noise across the graph, which would impact the performance and utility of the GNN. The LPGNN process itself is not altered, but we verify the impact of this attack on the trained GNN.}
 \caption{Visualization of Node Injection Attack on LPGNN.}    \label{fig:nodeinjectionattackoverview}
\end{figure*}

We first decide the number of nodes $n$ that the adversary is going to inject. Next we identify the top $n$ nodes from the graph that have the highest degree (number of directly connected neighbors). The adversary is going to create a node for each of these $n$ nodes and form an edge between them. We visually describe  this attack in Figure \ref{fig:nodeinjectionattackoverview}.
Since this is a locally private framework for GNNs, we assume a black box setting for the attack, which means that the nodes are crafted without much knowledge about the features and labels of the nodes they will be connected to. Hence, these values for the injected nodes are going to be random.
The intuition behind connecting to the nodes with the highest degree is that during training of the graph neural network, the features and labels are going to be aggregated across all the neighbors. As a result, the noisy features and labels in the injected nodes are going to be mixed with those of the highest degree nodes. These, in turn, will spread the noise further away to all their neighbors in subsequent rounds. This way, we aim to maximize the impact of the attack with fewer node injections.
It is possible to increase the potential of this attack even more by assuming a white box attack, where the adversary has knowledge of the feature and label values of the nodes in the graph. With such knowledge, the adversary can carefully craft feature values and label values for the node to be injected into the graph. While this form of attack can be explored, in a locally private setting of GNNs, such a powerful adversary is likely to be impractical.

\textbf{Threat Model}: The attacker has the ability to add custom nodes at desired locations in the graph data. They also have the ability to create edges between the injected nodes and other existing nodes in the graph. Therefore, the attacker has knowledge about the graph architecture but does not know the data contained within the nodes. Since the node data is unknown to the attacker, we consider this a black-box attack.
% \textbf{Threat Model}: 
The adversary in this attack aims to target the architecture of the graph, namely the nodes in the graph. The aim is to maliciously add nodes to the existing graph, so that the trained GNN would be misled to make wrong predictions. The disruption caused by this attack is both of utility and accuracy.

\subsubsection{\textbf{Label-Flipping Attack}}
\label{subsubsec:label_flipping_attack}

\begin{figure*}[t]
    \centering
    \includegraphics[width=1\linewidth]{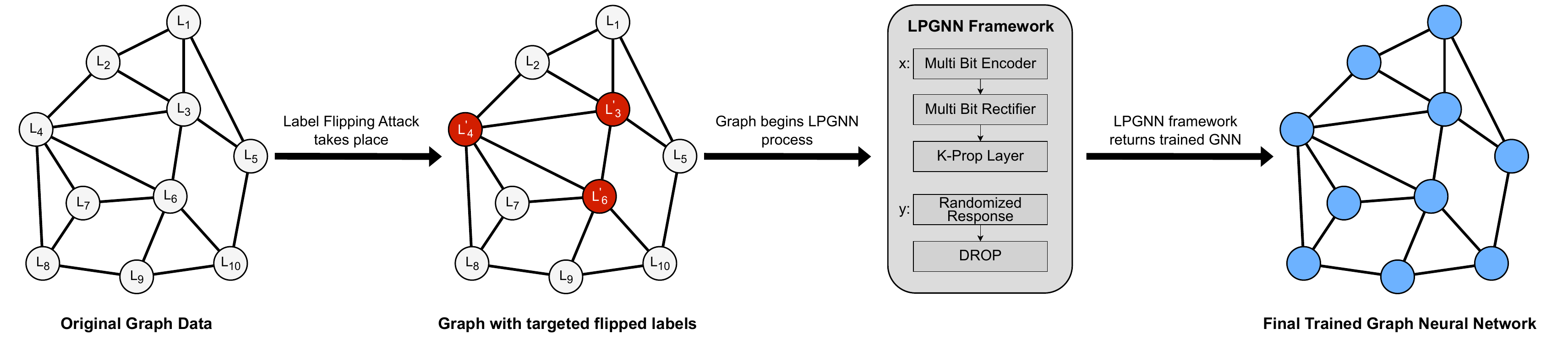}
    \captionsetup{justification=centering}
    % \caption{Visualization of Label Flipping Attack on LPGNN: There are $n$ nodes in the graph, each of which holds sensitive data features and labels. The attack can target nodes based on criteria such as highest node degree, flipping the data labels of targeted nodes. This aims to mislead and spread wrong correlation between features and labels across the graph, which would impact the performance and utility of the GNN. The LPGNN process itself is not altered, but we verify the impact of this attack on the trained GNN.}
        \caption{Visualization of Label Flipping Attack on LPGNN.}
    \label{fig:labelflippingattackoverview}
\end{figure*}

% \hfill \vspace{0.5mm} \break
The next form of attack we design against LPGNN is label-flipping, in which the adversary has the power to modify the label values of compromised nodes. Although this model of attack is already established \cite{advLabelFlippingAttacknDefence, multiLabelFlippingAdvAttackOnGNN}, we still aim to assess its performance in the context of LPGNNs. The attack is carried out as follows. (i) The most lucrative set of nodes from the graph that would act as the compromised nodes is first identified. This selection can happen based on metrics such as node-degree, which was used in the previous attack. Other suitable metrics may also be chosen. (ii) Next, the adversary randomly flips the values of labels in these nodes to some other label value. (iii) The LPGNN framework starts after this, without any changes to it.
It is important to note that the attack occurs before the application of LDP in LPGNN. Hence the attack is carried out prior to the LPGNN starting point and not during the LPGNN process itself. The attack has been visually described in Figure \ref{fig:labelflippingattackoverview} for ease of understanding.
%This is how it is different from the other ones.

\textbf{Threat Model}: The attacker has the ability to access and modify the node labels in the graph. Therefore, the attacker has knowledge about the label values for nodes in the graph but does not know the node feature values. Since the node data is known to the attacker, we consider this to be a white-box attack.
% \textbf{Threat Model}: 
The adversary in this attack aims to target the values of node labels in the graph. The aim is to maliciously flip the label values, so that the GNN trained on these changed label values would be misled to make wrong predictions. The disruption caused by this attack is both of utility and accuracy.

% \textbf{For our attacks, we picked the nodes with the highest degree as the compromised nodes. The results for our experiments can be found in section.
% }
\subsection{Inference Attack}
\label{subsec:gnn_inference_attack}

\begin{figure*}[t]
    \centering
    \includegraphics[width=1\linewidth]{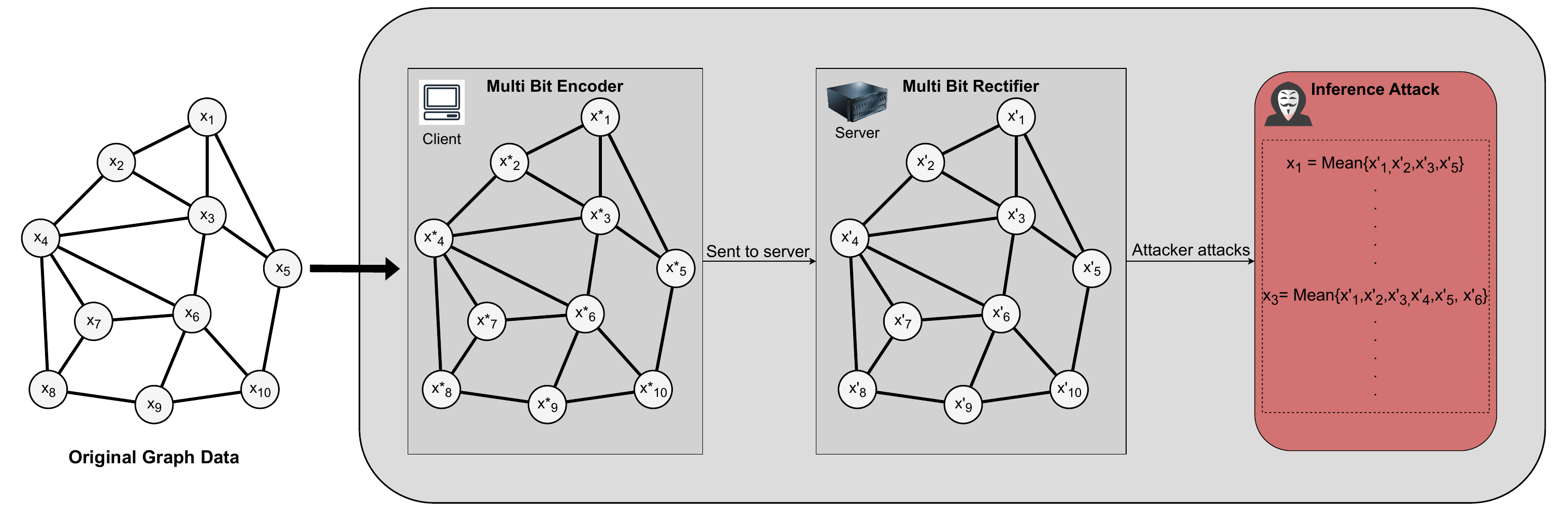}
    \captionsetup{justification=centering}
    % \caption{Visualization of Inference Attack on LPGNN: There are $n$ nodes in the graph, each of which holds sensitive data features. The attack aims to target and infer the sensitive data values of the nodes by collective aggregation of noisy values from a node's neighborhood. The consequence is breaking of the privacy guarantees offered by LDP. The attack is conducted during the LPGNN process, after the application of multi-bit rectifier.}
     \caption{Visualization of Inference Attack on LPGNN.}
    \label{fig:inferenceattackoverview}
\end{figure*}

We now look at a specific form of inference attack that was developed by us against LPGNN. The attack is based on an understanding of how graph data is generally structured - nodes with similar data are closer to each other and connected by an edge. Leveraging this information, we do an analysis of the aggregated values, inferring what possibly was the original values of the features. Unlike the previous two, this attack takes place during the process of LPGNN, right after the LDP mechanism is applied and the noisy feature vector is sent to the server. The steps for the attack are delineated below. (i) We first identify the nodes with large degrees in the graph. These nodes are theoretically more susceptible and weaker against attacks such as ours, since it depends on the aggregated values from the neighbors. (ii) The first step of LPGNN is allowed to happen - the application of Multi-Bit Encoder as the LDP mechanism for the feature vector. The outcome of this is biased, but it is sent to the server. (iii) On the server, the Multi-Bit Rectifier is applied, which brings the bias introduced into feature values to zero. (iv) The adversary on the server identifies the noisy features that belong to the targeted nodes and all of its direct neighbors. Then it performs the mean aggregation of the responses, neighborhood wise, forming the mean vector. (v) We conclude that for each dimension in the mean feature vector, its value is the predicted original value of the corresponding feature on targeted node.

Since the noisy values of the features carry zero bias from the application of LDP mechanism, if the neighborhood really holds similar feature values, we can expect their mean to converge to the original value of the feature. It is also important to note that these steps happen simultaneously for all the features belonging to the targeted node as shown in Figure \ref{fig:inferenceattackoverview}.
% So, if a node has a feature vector of dimension $m$, the attack is happening independently, but simultaneously, on all the $m$ features.
To measure the success of our attack, we plan to use two metrics - cosine similarity between the predicted feature vector and the original feature vector, and mean feature difference between the values of the predicted feature vector and the original feature vector. Since our attack design is based on an assumption, we conducted experiments to verify its credibility and applicability as outlined in Section \ref{sec:experiments}.

\textbf{Threat Model}: The attacker has the ability to access and read the noisy node feature and label data during the process of LPGNN. This can be achieved by compromising the server that oversees the LPGNN framework process. Since the attacker would be only getting access to the noisy data after the application of the LDP algorithm, we consider this to be a black-box attack.
% \textbf{Threat Model}: 
The adversary aims to infer the original values of the node data from the noisy values sent to the aggregation server. The aim is to combine and process the noisy data, so that the original values of the node data can be recovered. The disruption caused by this attack is both of privacy and integrity of LPGNN framework.

\subsection{Poisoning Attack}
\label{subsec:poisoningattack}

% It was observed that although our inference attack was founded on a intuitively meaningful assumption, the results are not very strong (Section \ref{sec:experiments}). 
We further looked at a different form of attack, where the adversary has the power to poison the targeted graph nodes and inference attacks then carried out by the server. Note that the inference attack did not involve any form of data poisoning. In this setting of the attack, 
% we assume that we are dealing with graph data, whose nodes are comprised of feature vectors constituting binary features. 
% That is, the private features in the graph data are binary in nature, with values denoting a \textit{"yes"} or a \textit{"no"} (usually stored as $1$ or $0$). 
we consider learning of the feature value or excluding a feature value as a possible original feature, as the violation of the privacy ($\epsilon$-LDP).
We put forth the design for this attack below.

\begin{figure*}[t]
    \centering
    \includegraphics[width=1\linewidth]{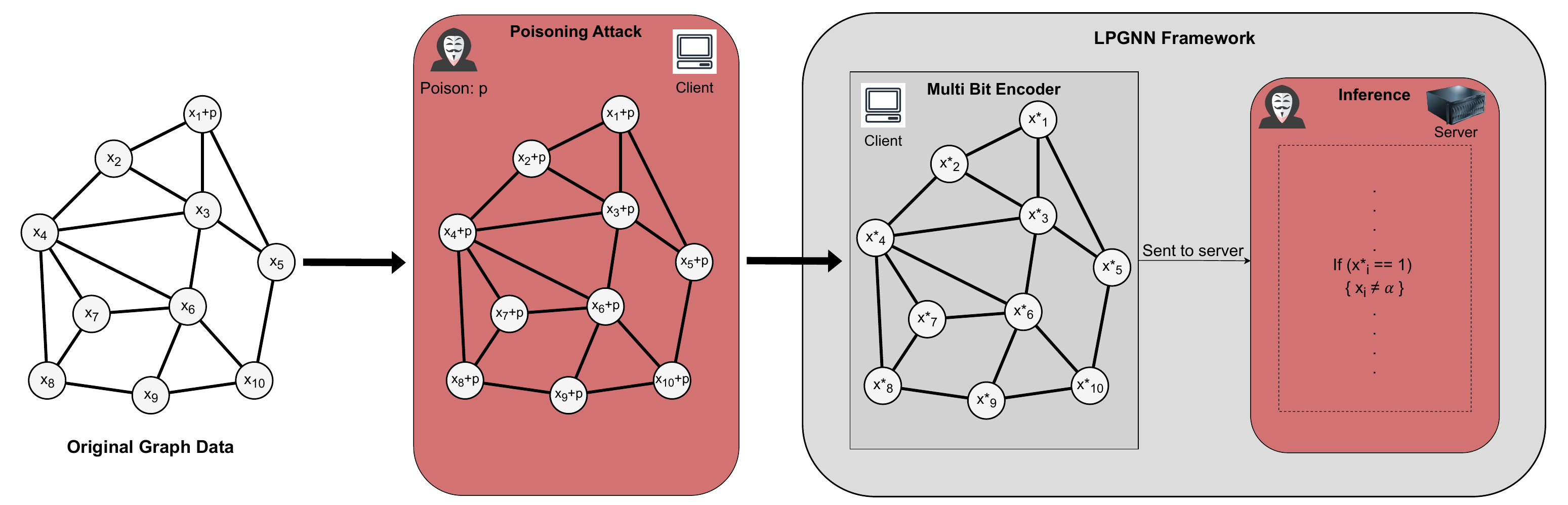}
    \captionsetup{justification=centering}
    % \caption{Visualization of Poisoning Attack on LPGNN: There are $n$ nodes in the graph, each of which holds sensitive data features. The attack aims to target and infer the sensitive data feature values of the nodes by injecting a poison before LPGNN starts, the attacker then infers the original private values from noisy values received at server. The consequence is breaking of the privacy guarantees offered by LDP. The attack is conducted both before and during the LPGNN process, specifically after the application of multi-bit encoder.}
        \caption{Visualization of Poisoning Attack on LPGNN.}
    \label{fig:poisoningattackoverview}
\end{figure*}

\textbf{Threat Model}: The attacker has the ability to add pre-computed poison to the targeted nodes in the graph data. They, however, do not have access to the original data values and is only able to add this poison to the data. This means that the raw feature values of the nodes are protected from direct access by the adversary and hence remain private. The attacker has the knowledge about the agreed-upon parameters for the LPGNN training process, such as $\alpha$, $\beta$, $\epsilon_x$, $m$, etc. 
The attacker also has knowledge about what values are used to store the binary representations of the features in the feature vector (for example, $1$ and $0$). The server carrying out the inference attack has the knowledge of which nodes have been compromised through poisoning by the attacker. Thus, this is a gray-box attack.

% \textbf{Threat Model}: 
The adversary in this attack specifically targets the private features of graph nodes. Their aim is to break the privacy guarantee offered by the LDP mechanism, which is the multi-bit mechanism. It attempts to infer additional information about the feature values, which otherwise could not be inferred with certainty due to the noise injected by LDP. The disruption caused by this attack is both of confidentiality and integrity. With the injected poison, the attacker can infer the original private feature values and also in the process degrade the utility and performance of the trained LPGNN.

\begin{algorithm}
\caption{Multi-Bit Encoder~\cite{lpgnn}}
\label{algorithm:mbm}
\begin{algorithmic}[1]
    \Statex \textbf{Input:} feature vector $x \in [\alpha, \beta]^d$, privacy budget $\varepsilon > 0$; range parameters $\alpha$ and $\beta$; sampling parameter $m \in \{1, 2, \ldots, d\}$.
    \Statex \textbf{Output:} encoded vector $x^* \in \{-1, 0, 1\}^d$.
\State Let $S$ be a set of $m$ values drawn uniformly at random without replacement from $\{1, 2, \ldots, d\}$
\For{$i \in \{1, 2, \ldots, d\}$}
    \State $s_i = 1$ if $i \in S$ otherwise $s_i = 0$
    \State $t_i \sim \text{Bernoulli}\left(\frac{1}{e^{\varepsilon/m} + 1} + \frac{x_i - \alpha}{\beta - \alpha} \cdot \frac{e^{\varepsilon/m} - 1}{e^{\varepsilon/m} + 1}\right)$
    \State $x^*_i = s_i \cdot (2t_i - 1)$
\EndFor
\State \Return $x^* = [x^*_1, \ldots, x^*_d]^T$
\end{algorithmic}
\end{algorithm}

Algorithm \ref{algorithm:mbm} outlines the multi-bit encoder mechanism used in the LDP process of LPGNN~\cite{lpgnn}. As seen in the algorithm, the randomness in this mechanism primarily comes from the Bernoulli trial it conducts based on a value computed using the raw feature value and various parameters. Our proposed data-poisoning attack particularly targets Line 4 of the algorithm.

% \begin{equation}
%     p = \left(\frac{1}{e^{\varepsilon/m} + 1} + \frac{x_i - \alpha}{\beta - \alpha} \cdot \frac{e^{\varepsilon/m} - 1}{e^{\varepsilon/m} + 1}\right)
%     \label{eq:poisonprobability}
% \end{equation}

We now mention the preliminary conditions for an attacker to best conduct this attack. The feature values should preferably be not of contiguous range, but from a set of finite discrete values in the domain (binary being the most favorable). When the feature values are not discrete, the attack turns into a membership-exclusion attack, still violating LDP guarantees. The attacker has knowledge about how the feature values are stored (what a particular value of feature represents). They also have the power to anonymously inject poison into the features before the LPGNN process starts. When these conditions are met, the adversary can successfully carry out the inference attack on poisoned nodes. Let us assume that one of the feature values is $\alpha$ itself. Then, the poison \textit{p} that is going to be injected into the features is given by,

\begin{equation}
    \text{p} = \frac{\alpha-\beta}{e^{\epsilon/m}-1}
    \label{eq:poison}
\end{equation}

\pagestyle{plain}

This poison is added to all the feature values of the nodes targeted.
The attacker is able to infer some of the private feature values when this poison is injected, as outlined by the following proof.
% Once the features are poisoned, the probability that the Bernoulli trial in the algorithm is going to output $1$ is zero. This implies that, for original $x_i=\alpha$, the output from the Bernoulli trial is always going to be $0$. Therefore, when the server sees a $1$ in the noisy feature vector, it can immediately conclude that the particular feature certainly did not hold $\alpha$ as the feature value. This clearly violates the guarantee of Local Differential Privacy, as LDP bounds the ratio of probability of such conclusions within a parameter, also known as privacy budget $\epsilon$.

% In binary feature scenarios, the inference is much stronger, as the server can immediately conclude that the original feature value was the value other than $\alpha$. Therefore, by virtue of the data poisoning, the adversary on the server is able to conduct successful inference attacks, clearly violating the node privacy and LDP guarantees. 
% The proof for our attack is given in the proofs section and the results for experiments conducted simulating these attacks can be found in Section \ref{subsec:data_poisoning_results}.

% The attacker is able to infer some of the private feature values when the feature vectors are injected with a poison of value $p = -\frac{\beta - \alpha}{e^{\epsilon/m}-1}$ without the node's knowledge.

\label{proof:poisoning_attack}
Let us consider the basic LPGNN setting, where there is no poisoning and the features undergo feature-perturbation through the multi-bit encoder and the multi-bit rectifier. 
% The algorithm for multi-bit encoder is given here - \ref{algorithm:mbm}. 
For values that are picked at random from the set of values ${1, 2, ..., d}$, the output depends on the Bernoulli draw made using the probability shown.

According to Algorithm 1, for any dimension \( i \in \{1,2,\ldots,d\} \), it can be seen that \( x^*_i \in \{-1,0,1\} \). The case \( x^*_i = 0 \) occurs when \( i \notin S \) with probability \( 1 - \frac{m}{d} \). 
% Therefore,
% \begin{equation}
% \Pr[M(x_1)_i = 0] = 1 - \frac{m}{d}
% \end{equation}
% \begin{equation}
% \Pr[M(x_2)_i = 0] = 1 - \frac{m}{d}
% \end{equation}
In the case of \( x^*_i = \{-1,1\} \), the probability of getting \( x^*_i = 1 \) ranges from \( \frac{m}{d} \cdot \frac{1}{e^{\varepsilon/m}+1} \) to \( \frac{m}{d} \cdot \frac{e^{\varepsilon/m}}{e^{\varepsilon/m}+1} \) depending on the value of \( x_i \). Analogously, the probability of \( x^*_i = -1 \) also varies from \( \frac{m}{d} \cdot \frac{1}{e^{\varepsilon/m}+1} \) to \( \frac{m}{d} \cdot \frac{e^{\varepsilon/m}}{e^{\varepsilon/m}+1} \). Therefore,

\begin{equation}
\frac{\Pr[M(x_1)_i \in \{-1,1\}]}{\Pr[M(x_2)_i \in \{-1,1\}]} \leq \frac{\max \Pr[M(x_1)_i \in \{-1,1\}]}{\min \Pr[M(x_2)_i \in \{-1,1\}]}
\end{equation}

The above ratio turns out to be $e^{\epsilon/m}$ which multiplied across the $m$ picked features, leads to $e^\epsilon$, and hence the LDP guarantee. Now consider the following setting where $p$ is added to all the feature values by an attacker before the LDP algorithm is invoked.
% We begin under the assumption that the feature values are discrete and come from a set of possible values between $\alpha$ and $\beta$, with one of the feature values being represented by $\alpha$.
Recollect the expression that is being used to determine the probability for the Bernoulli trial in Algorithm \ref{algorithm:mbm}. It is given by

\begin{equation}
    \text{probability} = \left(\frac{1}{e^{\varepsilon/m} + 1} + \frac{x_i - \alpha}{\beta - \alpha} \cdot \frac{e^{\varepsilon/m} - 1}{e^{\varepsilon/m} + 1}\right)
    \label{eq:prob}
\end{equation}

The new feature values after the poison is added are given by $x_i^* = x_i + \frac{\alpha-\beta}{e^{\epsilon/m}-1}$
% \begin{equation}
%     x_i^* = x_i + \frac{\alpha-\beta}{e^{\epsilon/m}-1}
%     \label{eq:xi}
% \end{equation}
We now consider the case where $x_i$ is equal to $\alpha$. The probability expression condenses to

\begin{equation}
    \begin{split}
        \text{probability} &= \frac{1}{e^{\epsilon/m}+1} + \frac{\alpha + p - \alpha}{\beta - \alpha} \cdot \frac{e^{\epsilon/m}-1}{e^{\epsilon/m}+1}\\
        % &= \frac{1}{e^{\epsilon/m}+1} - \frac{1}{e^{\epsilon/m}-1} \cdot \frac{e^{\epsilon/m}-1}{e^{\epsilon/m}+1}\\
        &= \frac{1}{e^{\epsilon/m}+1} - \frac{1}{e^{\epsilon/m}+1} 
        = 0
    \end{split}
    \label{eq:prob0}
\end{equation}

Therefore, the outcome of the Bernoulli trial will always be $0$ when the original feature value $x_i=\alpha$. Hence, whenever the server receives $1$ as the noisy feature value, it can immediately conclude that the original value of the feature must be something other than $\alpha$, thereby inferring sensitive information about the features. This clearly violates the guarantee of Local Differential Privacy, as LDP bounds the ratio of probability of such conclusions within a parameter, i.e., the privacy budget $\epsilon$.
In a binary feature value setting, the original feature value is immediately leaked. Even if the feature values are stored such that they begin at a value larger than $\alpha$, this can be overturned by subtracting its difference with $\alpha$ from the poison to be added.
This completes the proof that injecting poison $p$ into features will result in successful inference attacks. Figure \ref{fig:poisoningattackoverview} depicts an overview of this attack.

% \section{Analysis of Attacks}

% In this section, we shall thoroughly analyze the consequences of our attacks, in particular, the privacy analysis of the attacks. We show how our poisoning attack breaks the privacy guarantee offered by the local differential private algorithm used in the work \cite{lpgnn}. 

% \subsection{Privacy Analysis of Poisoning Attack}

% In section \ref{proof:poisoning_attack}, we showed how our poisoning attack could lead to successful inference attacks, thereby breaking privacy. In this section, we show how the LDP guarantee is being broken through the poisoning attack.

% Consider a private dataset with features

\section{Experiments}
\label{sec:experiments}

We simulated all the discussed attacks on the LPGNN framework across three graph neural network architectures, namely, GCN, GAT and SAGE. The experiments were done on four datasets $-$ Cora, PubMed, Facebook and LastFM, and the results provide a deeper understanding into the defense of LPGNN framework against the attacks. We begin with computing the baselines for our experiments, i.e., under no attack scenario.
For each dataset, we repeated the experiments using different GNN architectures. A feature privacy budget of $\epsilon_x = 8$ and a label privacy budget of $\epsilon_y = 4$ were used for the experiments. Table \ref{tab:baseline} illustrates the results obtained.

\begin{table}
\begin{center}
\caption{Baseline Results}
\label{tab:baseline}
\begin{tabular}{ |p{1.5cm}||p{1cm}|p{1cm}|p{1cm}|  }
 \hline
 \multicolumn{4}{|c|}{LPGNN Baselines (Accuracy in \%)} \\
 \hline
 Dataset & GCN & GAT & SAGE\\
 \hline
 Cora   & 79.9 & 56.3 & 77.4\\
 PubMed & 78.6 & 81 & 78.6\\
 Facebook & 85.3 & 57.1 & 83.9\\
 lastFM & 78.6 & 56.3 & 77.5\\
 \hline
\end{tabular}
\end{center}
  \end{table}
% \end{center}

\subsection{Classical Attack Experiments}
\label{subsec:classicalattackexp}

In this section, we present the results of the classical attacks introduced in Section \ref{subsec:classical_attacks}, simulated on the four datasets. We aim to understand the utility and applicability of existing attacks for this framework and conclude if such attacks are successful or not. We start with the Node Injection attacks.

\subsubsection{Node Injection Attack Results}
\label{subsubsec:node_injection_results}

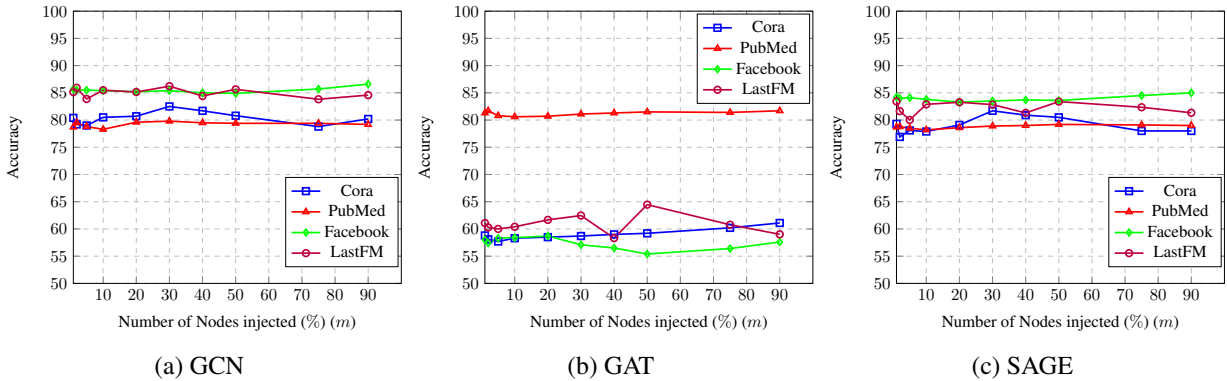
\begin{figure*}[t]
\advance\leftskip-0.00cm
 \resizebox{1\textwidth}{!}{
 \centering
\begin{subfigure}[b]{0.3\textwidth}
    \resizebox{\textwidth}{!}{
    \begin{tikzpicture}
    \begin{axis}[
        legend columns=1,
        xlabel={Number of Nodes injected ($\%$) ($m$)},
        ylabel={Accuracy},
        xmin=1, xmax=100,
        ymin=50, ymax=100,
        ytick={50,55,60,65,70,75,80,85,90,95,100},
        xtick={0,10,20,30,40,50,60,70,80,90},
        legend pos=south east,
        ymajorgrids=true,
        xmajorgrids=true,
        grid style=dashed,
        legend style={at={(axis cs:65,53)},anchor=south west},
    ]
    \addplot[
        color=blue,
        mark=square,
        line width=1.0pt
        ]
        coordinates {
        (1, 80.4) (2, 79.2) (5, 79) (10, 80.5) (20, 80.7) (30, 82.5) (40, 81.7) (50, 80.8) (75, 78.8) (90, 80.2)
        
        % (1, 0.0) (2, 36.0596) (3, 40.1042) (4, 42.8652) (5, 47.4892) (6, 47.2051) (7, 48.7518) (8, 51.6295) (9, 54.0228) (10, 52.3557) (11, 55.0768) (12, 58.7439) (13, 56.8394) (14, 58.0123) (15, 61.89) (16, 61.7411) (17, 62.1373) (18, 64.8402) (19, 63.2591) (20, 65.2547) (21, 63.6297) (22, 64.5008) (23, 63.8473) (24, 65.1172) (25, 67.7254) (26, 64.9198) (27, 68.5851) (28, 63.8065) (29, 66.9866) (30, 68.0064) (31, 67.5179) (32, 67.8251) (33, 72.7206) (34, 70.0717) (35, 68.5687) (36, 70.1286) (37, 69.1286) (38, 68.1273) (39, 71.7353) (40, 72.8199) (41, 72.578) (42, 74.6646) (43, 69.4947) (44, 75.06) (45, 70.578) (46, 71.3639) (47, 73.3948) (48, 74.1138) (49, 73.0923) (50, 75.2117)
        };
        \addlegendentry{Cora}

    \addplot[
        color=red,
        mark=triangle,
        line width=1.0pt
        ]
        coordinates {
        (1, 78.7) (2, 79.5) (5, 78.8) (10, 78.3) (20, 79.6) (30, 79.8) (40, 79.5) (50, 79.4) (75, 79.4) (90, 79.2)
        
         % (1, 0.0) (2, 37.3815) (3, 39.4006) (4, 42.673) (5, 43.8945) (6, 48.7955) (7, 53.1206) (8, 54.7179) (9, 56.8825) (10, 57.8072) (11, 55.5452) (12, 57.3121) (13, 57.7831) (14, 58.9609) (15, 61.5463) (16, 60.9736) (17, 62.074) (18, 63.3513) (19, 64.8056) (20, 64.2313) (21, 62.5561) (22, 65.4571) (23, 63.4066) (24, 62.0029) (25, 66.7342) (26, 66.5221) (27, 66.1627) (28, 68.0788) (29, 68.3731) (30, 67.9694) (31, 66.4963) (32, 67.9086) (33, 67.9529) (34, 70.5606) (35, 68.2496) (36, 71.2395) (37, 69.1575) (38, 70.6364) (39, 72.2102) (40, 71.2901) (41, 72.9067) (42, 67.7863) (43, 71.6934) (44, 75.7374) (45, 72.1164) (46, 72.3637) (47, 70.9857) (48, 69.6015) (49, 74.5148) (50, 76.6892)
        };
        \addlegendentry{PubMed}

    \addplot[
        color=green,
        mark=diamond,
        line width=1.0pt
        ]
        coordinates { 
                (1, 85.7) (2, 85.5) (5, 85.5) (10, 85.4) (20, 85.2) (30, 85.4) (40, 85) (50, 84.9) (75, 85.7) (90, 86.6)
                
         	% (1, 0.0) (2, 3.8891) (3, 4.3593) (4, 5.6519) (5, 6.2806) (6, 7.9) (7, 7.6266) (8, 7.5533) (9, 8.3063) (10, 7.9337) (11, 8.8484) (12, 10.8464) (13, 9.9513) (14, 10.0048) (15, 11.7684) (16, 10.1498) (17, 12.1013) (18, 10.6917) (19, 11.3792) (20, 12.5086) (21, 12.229) (22, 13.0426) (23, 13.5516) (24, 13.4597) (25, 14.8198) (26, 12.9984) (27, 15.4951) (28, 12.4397) (29, 14.6452) (30, 14.5066) (31, 14.3251) (32, 15.4714) (33, 15.6181) (34, 15.4615) (35, 16.4902) (36, 14.6119) (37, 16.0046) (38, 16.5072) (39, 15.3683) (40, 16.1762) (41, 15.8634) (42, 18.0795) (43, 16.3884) (44, 16.6456) (45, 16.4021) (46, 18.3389) (47, 16.5859) (48, 15.5356) (49, 17.3044) (50, 17.3529)
        };
        \addlegendentry{Facebook}
    \addplot[
        color=purple,
        mark=o,
        line width=1.0pt
        ]
        coordinates {
        (1, 85.1327) (2, 85.9289) (5, 83.8735) (10, 85.4828) (20, 85.1496) (30, 86.2168) (40, 84.4156) (50, 85.6409) (75, 83.8227) (90, 84.5793)
        };
        \addlegendentry{LastFM}

    \end{axis}
    \end{tikzpicture}}
     \caption{GCN} 
    \label{fig:gcn_node_injection_attack_results}
 \end{subfigure}
 \hfill
 \begin{subfigure}[b]{0.3\textwidth}
    \resizebox{\textwidth}{!}{
    \begin{tikzpicture}
    \begin{axis}[
        legend columns=1,
        xlabel={Number of Nodes injected ($\%$) ($m$)},
        ylabel={Accuracy},
        xmin=1, xmax=100,
        ymin=50, ymax=100,
        ytick={50,55,60,65,70,75,80,85,90,95,100},
        xtick={0,10,20,30,40,50,60,70,80,90},
        legend pos=south east,
        ymajorgrids=true,
        xmajorgrids=true,
        grid style=dashed,
        legend style={at={(axis cs:65,83)},anchor=south west},
    ]
    \addplot[
        color=blue,
        mark=square,
        line width=1.0pt
        ]
        coordinates {
        (1, 58.8) (2, 58.1) (5, 57.7) (10, 58.3) (20, 58.5) (30, 58.7) (40, 59) (50, 59.2) (75, 60.2) (90, 61.1)
        
        };
        \addlegendentry{Cora}

    \addplot[
        color=red,
        mark=triangle,
        line width=1.0pt
        ]
        coordinates {
        (1, 81.3) (2, 81.7) (5, 80.8) (10, 80.6) (20, 80.7) (30, 81.1) (40, 81.3) (50, 81.5) (75, 81.4) (90, 81.7)
        
        };
        \addlegendentry{PubMed}

    \addplot[
        color=green,
        mark=diamond,
        line width=1.0pt
        ]
        coordinates {
                (1, 58) (2, 57.4) (5, 58.3) (10, 58.4) (20, 58.7) (30, 57.1) (40, 56.5) (50, 55.4) (75, 56.4) (90, 57.6)
                
        };
        \addlegendentry{Facebook}
    \addplot[
        color=purple,
        mark=o,
        line width=1.0pt
        ]
       coordinates {
        (1, 61.0898) (2, 60.288) (5, 60) (10, 60.4009) (20, 61.6714) (30, 62.4619) (40, 58.323) (50, 64.4664) (75, 60.7736) (90, 59.0175)
        };
        \addlegendentry{LastFM}

    \end{axis}
    \end{tikzpicture}}
     \caption{GAT} 
    \label{fig:gat_node_injection_attack_results}
 \end{subfigure}
 \hfill
 \begin{subfigure}[b]{0.3\textwidth}
    \resizebox{\textwidth}{!}{
    \begin{tikzpicture}
    \begin{axis}[
        legend columns=1,
        xlabel={Number of Nodes injected ($\%$) ($m$)},
        ylabel={Accuracy},
        xmin=1, xmax=100,
        ymin=50, ymax=100,
        ytick={50,55,60,65,70,75,80,85,90,95,100},
        xtick={0,10,20,30,40,50,60,70,80,90},
        legend pos=south east,
        ymajorgrids=true,
        xmajorgrids=true,
        grid style=dashed,
        legend style={at={(axis cs:65,53)},anchor=south west},
    ]
    \addplot[
        color=blue,
        mark=square,
        line width=1.0pt
        ]
        coordinates {
        (1, 79.3) (2, 76.9) (5, 78.1) (10, 77.9) (20, 79.1) (30, 81.7) (40, 80.9) (50, 80.5) (75, 78) (90, 78)
        
        };
        \addlegendentry{Cora}

    \addplot[
        color=red,
        mark=triangle,
        line width=1.0pt
        ]
        coordinates {
        (1, 78.7) (2, 78.8) (5, 78.5) (10, 78.2) (20,78.6) (30, 78.9) (40, 79) (50, 79.2) (75, 79.1) (90, 79)
        
        };
        \addlegendentry{PubMed}

    \addplot[
        color=green,
        mark=diamond,
        line width=1.0pt
        ]
        coordinates {
                (1, 84.2) (2, 84) (5, 84.1) (10, 83.8) (20, 83.3) (30, 83.5) (40, 83.7) (50, 83.6) (75, 84.5) (90, 85)
                
        };
        \addlegendentry{Facebook}
    \addplot[
        color=purple,
        mark=o,
        line width=1.0pt
        ]
       coordinates {
        (1, 83.4218) (2, 81.6431) (5, 80.0282) (10, 82.8854) (20, 83.2693) (30, 82.8346) (40, 81.31) (50, 83.3936) (75, 82.349) (90, 81.3382)
        };
        \addlegendentry{LastFM}

    \end{axis}
    \end{tikzpicture}}
 
    \caption{SAGE} 
    \label{fig:sage_node_injection_attack_results}
 \end{subfigure}
 \hfill
 }
 \caption{Node Injection Attack results against the number of nodes injected into the graph as a percentage of total initial nodes}
 \label{fig:injection_attack_results}
\end{figure*}

We carried out node injection attacks on all four datasets using the three GNN architectures independently. The attack was simulated at varying degrees
% , beginning with a very small scale injection attack, where only an extra $1\%$ amount of random nodes are injected into the graph and go upto a massive scale attack. 
 of scale. Such attacks prioritize connecting new nodes with existing nodes of the highest degree, as described in Section \ref{subsubsec:node_injection_attack}. The results for the attacks for varying number of nodes injected are shown in Figures \ref{fig:injection_attack_results}(a)-(c). From the figures, it is seen that the accuracy of the GNN does not vary much even after conducting a wide scale node injection attack. This implies that the node injection attack is not particularly successful in the context of the locally private GNNs. The main reason can be attributed to the fact that since this is a black box attack, crafting of the nodes is random, which does not fully utilize the potential of the attack. The attack can be more successful by making the adversary aware of everything in the context where the node is being injected, but in a locally private setting, it is impractical to assume that for an adversary.
It is also important to note that we are trying to mislead the GNN, which is already misled due to the noise introduced by the LDP mechanism. Trying to propagate random features and labels through node injection, therefore, has limited utility, because it could also lead a misled GNN in certain aspects correctly to the right prediction.

% below is acc vs e_x
\begin{figure*}[t]
\advance\leftskip-0.00cm
 \resizebox{1\textwidth}{!}{
 \centering
\begin{subfigure}[b]{0.3\textwidth}
    \resizebox{\textwidth}{!}{
    \begin{tikzpicture}
    \begin{axis}[
        legend columns=1,
        xlabel={feature privacy budget ($\epsilon_x$)},
        ylabel={Accuracy},
        xmin=1, xmax=100,
        ymin=50, ymax=100,
        ytick={50,55,60,65,70,75,80,85,90,95,100},
        xtick={0,5,10,20,30,40,50,60,70,80,90,100},
        legend pos=south east,
        ymajorgrids=true,
        xmajorgrids=true,
        grid style=dashed,
        legend style={at={(axis cs:65,53)},anchor=south west},
    ]
    \addplot[
        color=blue,
        mark=square,
        line width=1.0pt
        ]
        coordinates {
        (1, 75.1699) (2, 76.647) (3, 78.6558) (4, 78.8331) (5, 79.9557) (7, 80.4874) (10, 80.5318) (20, 80.8272) (30, 81.3589) (40, 81.0044) (50, 81.0783) (75, 80.9601) (100, 81.0487)
        };
        \addlegendentry{Cora}

    \addplot[
        color=red,
        mark=triangle,
        line width=1.0pt
        ]
        coordinates {
        (1, 78.2126) (2, 78.4216) (3, 78.4764) (4, 78.5048) (5, 78.2816) (7, 78.4277) (10, 78.3059) (20, 77.815) (30, 77.5309) (40, 77.3747) (50, 77.3321) (75, 76.9) (100, 76.6261)

        };
        \addlegendentry{PubMed}

    \addplot[
        color=green,
        mark=diamond,
        line width=1.0pt
        ]
        coordinates {
               (1, 82.4279) (2, 83.8074) (3, 84.167) (4, 84.5924) (5, 85.0374) (7, 85.0481) (10, 85.4201) (20, 85.6497) (30, 85.7707) (40, 85.8936) (50, 85.9256) (75, 85.9701) (100, 86.0182)

        };
        \addlegendentry{Facebook}
    \addplot[
        color=purple,
        mark=o,
        line width=1.0pt
        ]
        coordinates {
        (1, 72.4167) (2, 75.8272) (3, 76.3354) (4, 78.5997) (5, 79.2829) (7, 79.8363) (10, 81.8746) (20, 83.45) (30, 84.9068) (40, 85.1214) (50, 85.6465) (75, 85.8385) (100, 85.8216)

        };
        \addlegendentry{LastFM}

    \end{axis}
    \end{tikzpicture}}
     \caption{GCN} 
    \label{fig:gcn_node_injection_attack_eps_results}
 \end{subfigure}
 \hfill
 \begin{subfigure}[b]{0.3\textwidth}
    \resizebox{\textwidth}{!}{
    \begin{tikzpicture}
    \begin{axis}[
        legend columns=1,
        xlabel={feature privacy budget ($\epsilon_x$)},
        ylabel={Accuracy},
        xmin=1, xmax=100,
        ymin=50, ymax=100,
        ytick={50,55,60,65,70,75,80,85,90,95,100},
        xtick={0,5,10,20,30,40,50,60,70,80,90,100},
        legend pos=south east,
        ymajorgrids=true,
        xmajorgrids=true,
        grid style=dashed,
        legend style={at={(axis cs:30,83)},anchor=south west},
    ]
    \addplot[
        color=blue,
        mark=square,
        line width=1.0pt
        ]
        coordinates {
        (1, 49.1137) (2, 50.3102) (3, 53.2939) (4, 54.195) (5, 54.8154) (7, 55.7903) (10, 59.6603) (20, 66.4845) (30, 71.8907) (40, 77.6219) (50, 80.2511) (75, 81.6691) (100, 81.7282)

        };
        \addlegendentry{Cora}

    \addplot[
        color=red,
        mark=triangle,
        line width=1.0pt
        ]
        coordinates {
        (1, 60.3246) (2, 65.2992) (3, 74.7292) (4, 80.6492) (5, 80.7121) (7, 80.6918) (10, 80.5376) (20, 80.1704) (30, 79.8945) (40, 79.5597) (50, 79.3974) (75, 78.8639) (100, 78.2349)

        };
        \addlegendentry{PubMed}

    \addplot[
        color=green,
        mark=diamond,
        line width=1.0pt
        ]
        coordinates {
               (1, 52.346) (2, 54.5301) (3, 54.7116) (4, 53.8661) (5, 53.9267) (7, 54.6013) (10, 58.2289) (20, 59.8434) (30, 64.1741) (40, 69.5141) (50, 75.7583) (75, 83.6294) (100, 86.125)
                
        };
        \addlegendentry{Facebook}
    \addplot[
        color=purple,
        mark=o,
        line width=1.0pt
        ]
       coordinates {
       (1, 51.9198) (2, 54.1446) (3, 56.2507) (4, 54.4777) (5, 54.5059) (7, 57.4139) (10, 58.6222) (20, 58.515) (30, 59.6047) (40, 61.9819) (50, 61.4003) (75, 64.0373) (100, 65.8893)
        };
        \addlegendentry{LastFM}

    \end{axis}
    \end{tikzpicture}}
     \caption{GAT} 
    \label{fig:gat_node_injection_attack_eps_results}
 \end{subfigure}
 \hfill
 \begin{subfigure}[b]{0.3\textwidth}
    \resizebox{\textwidth}{!}{
    \begin{tikzpicture}
    \begin{axis}[
        legend columns=1,
        xlabel={feature privacy budget ($\epsilon_x$)},
        ylabel={Accuracy},
        xmin=1, xmax=100,
        ymin=50, ymax=100,
        ytick={50,55,60,65,70,75,80,85,90,95,100},
        xtick={0,5,10,20,30,40,50,60,70,80,90,100},
        legend pos=south east,
        ymajorgrids=true,
        xmajorgrids=true,
        grid style=dashed,
        legend style={at={(axis cs:65,53)},anchor=south west},
    ]
    \addplot[
        color=blue,
        mark=square,
        line width=1.0pt
        ]
        coordinates {
        (1, 69.6307) (2, 71.6839) (3, 74.6381) (4, 76.4402) (5, 75.5687) (7, 76.4993) (10, 78.2866) (20, 79.1285) (30, 80.1329) (40, 80.1625) (50, 80.6499) (75, 81.0487) (100, 81.1521)
        };
        \addlegendentry{Cora}

    \addplot[
        color=red,
        mark=triangle,
        line width=1.0pt
        ]
        coordinates {
       (1, 76.5389) (2, 77.4234) (3, 77.6912) (4, 78.0564) (5, 77.9124) (7, 78.0645) (10, 78.1538) (20, 78.172) (30, 78.0037) (40, 77.9448) (50, 77.8921) (75, 77.7906) (100, 77.6547)
        
        };
        \addlegendentry{PubMed}

    \addplot[
        color=green,
        mark=diamond,
        line width=1.0pt
        ]
        coordinates {
                (1, 77.5027) (2, 80.1317) (3, 81.5219) (4, 82.4261) (5, 82.9708) (7, 83.5119) (10, 84.1865) (20, 85.1424) (30, 85.5287) (40, 85.4717) (50, 85.477) (75, 85.6835) (100, 85.7494)
        };
        \addlegendentry{Facebook}
    \addplot[
        color=purple,
        mark=o,
        line width=1.0pt
        ]
       coordinates {
       (1, 67.0582) (2, 72.6256) (3, 74.026) (4, 76.2168) (5, 77.3631) (7, 77.7244) (10, 79.5596) (20, 81.9706) (30, 81.9989) (40, 82.7273) (50, 82.9983) (75, 83.7324) (100, 83.5178)
        };
        \addlegendentry{LastFM}

    \end{axis}
    \end{tikzpicture}}
 
    \caption{SAGE} 
    \label{fig:sage_node_injection_attack_eps_results}
 \end{subfigure}
 \hfill
 }
 \caption{Node Injection Attack results against the feature privacy budget ($\epsilon_x$) where 10\% of total nodes are injected}
 \label{fig:injection_attack_eps_results}
\end{figure*}
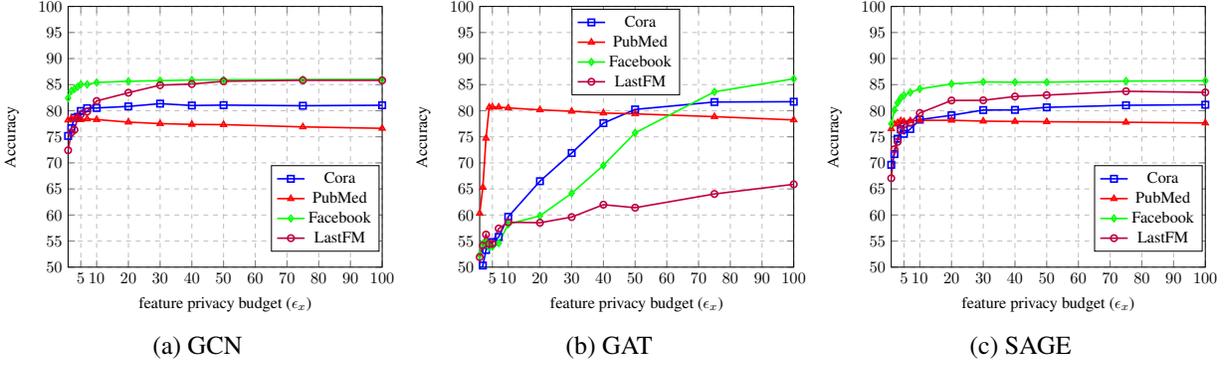

% \subsubsection{Impact of Feature Privacy Budget ($\epsilon_x$) under Node Injection}
% \label{subsubsec:node_injection_eps_results}

In addition to varying the percentage of injected nodes, we investigated the effect of the feature privacy budget ($\epsilon_x$) on the robustness of LPGNNs under a fixed node injection attack in Figures \ref{fig:injection_attack_eps_results}(a)-(c). For this experiment, we maintained a constant node injection rate of 10\% of the original graph size, varying the feature privacy budget $\epsilon_x$ across a range from 1 to 100, while keeping the label privacy budget constant. 
% The features and labels of the injected nodes were assigned randomly, simulating a black-box attack where the adversary lacks knowledge of neighbors' specifics. 
% The results of this experiment are depicted in Figure \ref{fig:injection_attack_eps_results}.
These figures reveal that increasing the feature privacy budget ($\epsilon_x$) means less noise is added to features. Less noise is expected to help the GNN perform better, improving accuracy even during an attack. The accuracy generally increases as $\epsilon_x$ gets larger, especially starting from low $\epsilon_x$ values. This was seen for most datasets and models tested. 
However, this improvement does not continue forever. Instead, the accuracy levels off at higher $\epsilon_x$ values (around 20-30). This plateau suggests that once feature noise is low enough, the main limit on performance becomes the effect of the attack itself and GNN limits, not the LDP noise.

\subsubsection{Label Flipping Attack Results}
\label{subsubsec:label_flipping_results}

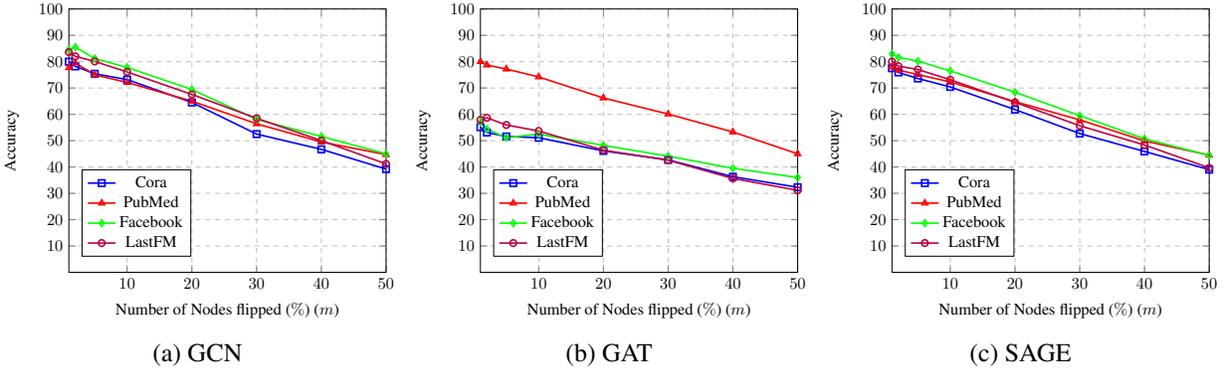
\begin{figure*}[t]
\advance\leftskip-0.00cm
 \resizebox{1\textwidth}{!}{
 \centering
\begin{subfigure}[b]{0.3\textwidth}
    \resizebox{\textwidth}{!}{
    \begin{tikzpicture}
    \begin{axis}[
        legend columns=1,
        xlabel={Number of Nodes flipped ($\%$) ($m$)},
        ylabel={Accuracy},
        xmin=1, xmax=50,
        ymin=0, ymax=100,
        ytick={10,20,30,40,50,60,70,80,90,100},
        xtick={0,10,20,30,40,50},
        legend pos=south east,
        ymajorgrids=true,
        xmajorgrids=true,
        grid style=dashed,
        legend style={at={(axis cs:3,6)},anchor=south west},
    ]
    \addplot[
        color=blue,
        mark=square,
        line width=1.0pt
        ]
        coordinates {
        (1, 80) (2, 78.2) (5, 75.4) (10, 73.2) (20, 64.5) (30, 52.5) (40, 46.7) (50, 39.2)
        
        };
        \addlegendentry{Cora}

    \addplot[
        color=red,
        mark=triangle,
        line width=1.0pt
        ]
        coordinates { %LATEST
        (1, 77.7) (2, 79.5) (5, 75) (10, 72.1) (20, 65) (30, 56.4) (40, 49.5) (50, 44.7)
        
        };
        \addlegendentry{PubMed}

    \addplot[
        color=green,
        mark=diamond,
        line width=1.0pt
        ]
        coordinates {
                (1, 84.5) (2, 85.5) (5, 81.2) (10, 77.8) (20, 69.4) (30, 58.1) (40, 51.6) (50, 45)
                
        };
        \addlegendentry{Facebook}
     \addplot[
        color=purple,
        mark=o,
        line width=1.0pt
        ]
       coordinates {
        (1, 83.6194) (2, 82.0215) (5, 80.0847) (10, 76.0982) (20, 67.5494) (30, 58.4924) (40, 50.0282) (50, 41.1688)
        };
        \addlegendentry{LastFM}

    \end{axis}
    \end{tikzpicture}}
     \caption{GCN} 
    \label{fig:gcn_label_flipping_attack_results}
 \end{subfigure}
 \hfill
 \begin{subfigure}[b]{0.3\textwidth}
    \resizebox{\textwidth}{!}{
    \begin{tikzpicture}
    \begin{axis}[
        legend columns=1,
        xlabel={Number of Nodes flipped ($\%$) ($m$)},
        ylabel={Accuracy},
        xmin=1, xmax=50,
        ymin=0, ymax=100,
        ytick={10,20,30,40,50,60,70,80,90,100},
        xtick={0,10,20,30,40,50},
        legend pos=south east,
        ymajorgrids=true,
        xmajorgrids=true,
        grid style=dashed,
        legend style={at={(axis cs:3,6)},anchor=south west},
    ]
    \addplot[
        color=blue,
        mark=square,
        line width=1.0pt
        ]
        coordinates {
        (1, 55.1) (2, 53.1) (5, 51.6) (10, 51.1) (20, 46.1) (30, 42.7) (40, 36.3) (50, 32.3)
        
        };
        \addlegendentry{Cora}

    \addplot[
        color=red,
        mark=triangle,
        line width=1.0pt
        ]
        coordinates {
        (1, 80) (2, 78.7) (5, 77.2) (10, 74.2) (20, 66.2) (30, 60.1) (40, 53.3) (50, 45)
        
        };
        \addlegendentry{PubMed}

    \addplot[
        color=green,
        mark=diamond,
        line width=1.0pt
        ]
        coordinates {
                (1, 57) (2, 54.4) (5, 51.1) (10, 52.5) (20, 48.2) (30, 44.1) (40, 39.5) (50, 36)
                
        };
        \addlegendentry{Facebook}
    \addplot[
        color=purple,
        mark=o,
        line width=1.0pt
        ]
       coordinates {
       (1, 57.8882) (2, 58.6957) (5, 55.9514) (10, 53.6872) (20, 46.3975) (30, 42.5635) (40, 35.6352) (50, 31.135) };
        \addlegendentry{LastFM}

    \end{axis}
    \end{tikzpicture}}
     \caption{GAT} 
    \label{fig:gat_label_flipping_attack_results}
 \end{subfigure}
 \hfill
 \begin{subfigure}[b]{0.3\textwidth}
    \resizebox{\textwidth}{!}{
    \begin{tikzpicture}
    \begin{axis}[
        legend columns=1,
        xlabel={Number of Nodes flipped ($\%$) ($m$)},
        ylabel={Accuracy},
        xmin=1, xmax=50,
        ymin=0, ymax=100,
        ytick={10,20,30,40,50,60,70,80,90,100},
        xtick={0,10,20,30,40,50},
        legend pos=south east,
        ymajorgrids=true,
        xmajorgrids=true,
        grid style=dashed,
        legend style={at={(axis cs:3,6)},anchor=south west},
    ]
    \addplot[
        color=blue,
        mark=square,
        line width=1.0pt
        ]
        coordinates {
        (1, 77.5) (2, 75.9) (5, 73.6) (10,70.4) (20, 61.8) (30, 52.7) (40, 45.9) (50, 39)
        
        };
        \addlegendentry{Cora}

    \addplot[
        color=red,
        mark=triangle,
        line width=1.0pt
        ]
        coordinates {
        (1, 78) (2, 76.8) (5, 75.1) (10, 72.3) (20, 64.8) (30, 57.9) (40, 49.9) (50, 44.5)
        
        };
        \addlegendentry{PubMed}

    \addplot[
        color=green,
        mark=diamond,
        line width=1.0pt
        ]
        coordinates {
                (1, 82.9) (2, 81.6) (5, 80.2) (10, 76.5) (20, 68.4) (30, 59.5) (40, 50.7) (50, 44.4)
                
        };
        \addlegendentry{Facebook}
    \addplot[
        color=purple,
        mark=o,
        line width=1.0pt
        ]
        coordinates {  
           (1, 80) (2, 78.2835) (5, 77.0243) (10, 73.1338) (20, 64.5455) (30, 55.6691) (40, 48.3512) (50, 39.6781)  
        };        
        \addlegendentry{LastFM}

    \end{axis}
    \end{tikzpicture}}
 
    \caption{SAGE} 
    \label{fig:sage_label_flipping_attack_results}
 \end{subfigure}
 \hfill
 % \caption{Experimental Results}
 % \label{fig:ExpResults}
 }
 \caption{Label Flipping Attack results against the number of node labels flipped in the graph (as a \% of total initial nodes)}
 \label{fig:label_flipping_attack_results}
\end{figure*}

Similar to the node injection attacks, we conducted Label Flipping attacks as explained in Section \ref{subsubsec:label_flipping_attack}, across four datasets and three different GNN architectures. Again we begin with a small scale label-flipping attack - only $1\%$ of the node labels being flipped, and we scale and test this attack upto $50\%$ of nodes being compromised. The results for this attack are shown in Figures \ref{fig:label_flipping_attack_results}(a)-(c). It is observed that the success of this label flipping attack increases linearly with the number of nodes impacted in the graph. Hence, we conclude that label flipping attacks are successful against locally private GNNs.
% and LPGNN framework is not immune to label flipping attacks, as with original GNNs.

% \subsubsection{Impact of Feature Privacy Budget ($\epsilon_x$) under Label Flipping}
% \label{subsubsec:label_flipping_eps_results}

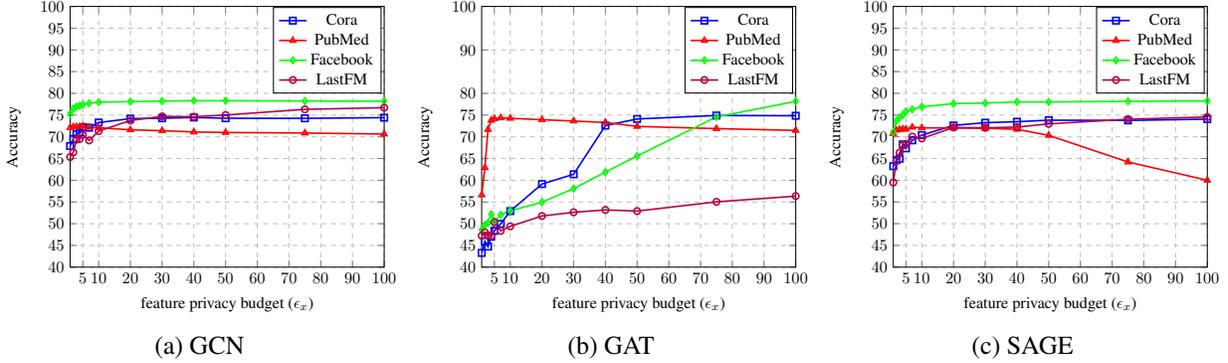
\begin{figure*}[t]
\advance\leftskip-0.00cm
 \resizebox{1\textwidth}{!}{
 \centering
\begin{subfigure}[b]{0.3\textwidth}
    \resizebox{\textwidth}{!}{
    \begin{tikzpicture}
    \begin{axis}[
        legend columns=1,
        xlabel={feature privacy budget ($\epsilon_x$)},
        ylabel={Accuracy},
        xmin=1, xmax=100,
        ymin=40, ymax=100,
        ytick={40,45,50,55,60,65,70,75,80,85,90,95,100},
        xtick={0,5,10,20,30,40,50,60,70,80,90,100},
        legend pos=south east,
        ymajorgrids=true,
        xmajorgrids=true,
        grid style=dashed,
        legend style={at={(axis cs:65,80)},anchor=south west},
    ]
    \addplot[
        color=blue,
        mark=square,
        line width=1.0pt
        ]
        coordinates {
        (1, 67.873) (2, 69.5126) (3, 70.6056) (4, 71.0487) (5, 72.2304) (7, 72.0827) (10, 73.2792) (20, 74.1802) (30, 74.2688) (40, 74.4461) (50, 74.2688) (75, 74.2541) (100, 74.4018)
        
        };
        \addlegendentry{Cora}

    \addplot[
        color=red,
        mark=triangle,
        line width=1.0pt
        ]
        coordinates {
        (1, 72.0937) (2, 72.2702) (3, 72.3189) (4, 72.3169) (5, 72.4325) (7, 72.1708) (10, 72.0917) (20, 71.6088) (30, 71.408) (40, 71.1057) (50, 70.9961) (75, 70.8663) (100, 70.633)
        
        };
        \addlegendentry{PubMed}

    \addplot[
        color=green,
        mark=diamond,
        line width=1.0pt
        ]
        coordinates {
                (1, 75.2278) (2, 76.4685) (3, 76.9722) (4, 77.1253) (5, 77.4333) (7, 77.7287) (10, 77.9619) (20, 78.0883) (30, 78.2182) (40, 78.3126) (50, 78.3126) (75, 78.2342) (100, 78.1808)

        };
        \addlegendentry{Facebook}
     \addplot[
        color=purple,
        mark=o,
        line width=1.0pt
        ]
       coordinates {
        (1, 65.3416) (2, 66.4257) (3, 69.3394) (4, 69.4918) (5, 70.4574) (7, 69.1756) (10, 71.2987) (20, 73.7549) (30, 74.7036) (40, 74.6132) (50, 75.0141) (75, 76.3185) (100, 76.6968)
        };
        \addlegendentry{LastFM}

    \end{axis}
    \end{tikzpicture}}
     \caption{GCN} 
    \label{fig:gcn_label_flipping_attack_eps_results}
 \end{subfigure}
 \hfill
 \begin{subfigure}[b]{0.3\textwidth}
    \resizebox{\textwidth}{!}{
    \begin{tikzpicture}
    \begin{axis}[
        legend columns=1,
        xlabel={feature privacy budget ($\epsilon_x$)},
        ylabel={Accuracy},
        xmin=1, xmax=100,
        ymin=40, ymax=100,
        ytick={40,45,50,55,60,65,70,75,80,90,100},
        xtick={5,10,20,30,40,50,60,70,80,90,100},
        legend pos=south east,
        ymajorgrids=true,
        xmajorgrids=true,
        grid style=dashed,
        legend style={at={(axis cs:65,80)},anchor=south west},
    ]
    \addplot[
        color=blue,
        mark=square,
        line width=1.0pt
        ]
        coordinates {
        (1, 43.2792) (2, 45.938) (3, 44.7267) (4, 47.0015) (5, 48.3604) (7, 49.9705) (10, 52.8951) (20, 59.1285) (30, 61.3737) (40, 72.5849) (50, 74.1064) (75, 74.904) (100, 74.8449)
        
        };
        \addlegendentry{Cora}

    \addplot[
        color=red,
        mark=triangle,
        line width=1.0pt
        ]
        coordinates {
        (1, 56.6383) (2, 62.8809) (3, 71.688) (4, 73.7979) (5, 74.157) (7, 74.3599) (10, 74.2605) (20, 73.9278) (30, 73.6397) (40, 73.2867) (50, 72.392) (75, 71.8888) (100, 71.4668)
        
        };
        \addlegendentry{PubMed}

    \addplot[
        color=green,
        mark=diamond,
        line width=1.0pt
        ]
        coordinates { 
                (1, 48.5867) (2, 49.7241) (3, 50.1833) (4, 52.1645) (5, 50.5287) (7, 52.0452) (10, 52.9619) (20, 54.9519) (30, 58.074) (40, 61.8815) (50, 65.5838) (75, 74.5746) (100, 78.1595)
                
        };
        \addlegendentry{Facebook}
    \addplot[
        color=purple,
        mark=o,
        line width=1.0pt
        ]
       coordinates {
       (1, 47.2388) (2, 48.0237) (3, 47.3123) (4, 47.2332) (5, 50.3557) (7, 48.3399) (10, 49.3958) (20, 51.7787) (30, 52.6256) (40, 53.1677) (50, 52.908) (75, 55.0311) (100, 56.3467) 
       };
        \addlegendentry{LastFM}

    \end{axis}
    \end{tikzpicture}}
     \caption{GAT} 
    \label{fig:gat_label_flipping_attack_eps_results}
 \end{subfigure}
 \hfill
 \begin{subfigure}[b]{0.3\textwidth}
    \resizebox{\textwidth}{!}{
    \begin{tikzpicture}
    \begin{axis}[
        legend columns=1,
        xlabel={feature privacy budget ($\epsilon_x$)},
        ylabel={Accuracy},
        xmin=1, xmax=100,
        ymin=40, ymax=100,
        ytick={40,45,50,55,60,65,70,75,80,85,90,95,100},
        xtick={0,5,10,20,30,40,50,60,70,80,90,100},
        legend pos=south east,
        ymajorgrids=true,
        xmajorgrids=true,
        grid style=dashed,
        legend style={at={(axis cs:65,80)},anchor=south west},
    ]
    \addplot[
        color=blue,
        mark=square,
        line width=1.0pt
        ]
        coordinates {
        (1, 63.2349) (2, 64.6233) (3, 64.9631) (4, 68.257) (5, 67.3412) (7, 69.1876) (10, 70.3988) (20, 72.6145) (30, 73.2496) (40, 73.4712) (50, 73.8257) (75, 73.7666) (100, 74.062)
        
        };
        \addlegendentry{Cora}

    \addplot[
        color=red,
        mark=triangle,
        line width=1.0pt
        ]
        coordinates {
        (1, 70.7669) (2, 71.5054) (3, 71.7083) (4, 71.7874) (5, 71.7488) (7, 72.2114) (10, 72.0532) (20, 72.0329) (30, 71.97) (40, 71.7265) (50, 70.3287) (75, 64.2017) (100, 59.998)
        
        };
        \addlegendentry{PubMed}

    \addplot[
        color=green,
        mark=diamond,
        line width=1.0pt
        ]
        coordinates { 
                (1, 71.1107) (2, 73.6258) (3, 74.4357) (4, 75) (5, 75.8366) (7, 76.3119) (10, 76.8993) (20, 77.6433) (30, 77.7625) (40, 78.0171) (50, 78.0153) (75, 78.1613) (100, 78.252)

        };
        \addlegendentry{Facebook}
    \addplot[
        color=purple,
        mark=o,
        line width=1.0pt
        ]
        coordinates {  
           (1, 59.4749) (2, 64.4495) (3, 66.3636) (4, 68.0802) (5, 68.3456) (7, 70.0621) (10, 69.616) (20, 72.14) (30, 72.0779) (40, 72.2304) (50, 72.9757) (75, 74.0429) (100, 74.5624) 
        };        
        \addlegendentry{LastFM}

    \end{axis}
    \end{tikzpicture}}
 
    \caption{SAGE} 
    \label{fig:sage_label_flipping_attack_eps_results}
 \end{subfigure}
 \hfill
 }
 \caption{Label Flipping Attack results against the feature privacy budget ($\epsilon_x$) where 10\% of total node labels are flipped}
 \label{fig:label_flipping_attack_eps_results}
\end{figure*}

To understand the interplay between feature privacy and label corruption, we study the effect of privacy budget ($\epsilon_x$) on the robustness of LPGNN under a fixed intensity label flipping attack. For this, we simulated an attack where labels of 10\% of nodes are flipped. These targeted nodes are chosen based on highest degrees. The true labels of these nodes were flipped to an incorrect label chosen randomly. During this attack scenario, we varied the feature privacy budget $\epsilon_x$ from 1 to 100, while the label privacy budget was kept constant (e.g., $\epsilon_y = 4$, consistent with other experiments). This allows observation of how the level of noise applied to features impacts the model's ability to handle incorrect label information propagated through the graph. The results
% , illustrating accuracy across different $\epsilon_x$ values for various GNN architectures under this attack, 
are depicted in Figures~\ref{fig:label_flipping_attack_eps_results}(a)-(c).\\

% x_eps attack below

\subsubsection{Impact of Feature and Label Perturbation Step Sizes}
\label{subsubsec:resultsonperturbationstepsize}
For the next set of experiments assessing the impact of step sizes on accuracy in the context of node injection or label flipping attacks, we followed a consistent metric of 10\%, 30\%, and 50\% of the total number of nodes being attacked/injected. All experiments were carried out using the GCN architecture. The feature privacy budget ($\epsilon_x$) was kept constant at 1 across all settings to isolate the impact of the hyperparameter under focus.

% \subsubsection{Impact of Feature Perturbation Step Size ($x_{\text{step}}$) under Node Injection Attack} \label{subsubsec:xstep_node_injection}

To assess the influence of feature perturbation intensity, we conducted experiments under a node injection attack scenario. Specifically, we injected new nodes into the graph at varying rates: 10\%, 30\%, and 50\% of the total number of nodes. 
The injected nodes were connected to existing high-degree nodes and assigned feature vectors that were perturbed using different values of the feature perturbation step size ($x_{\text{step}}$). All experiments were carried out using the GCN architecture. The feature privacy budget ($\epsilon_x$) was kept constant at 1 across all settings to isolate the impact of $x_{\text{step}}$. 
This evaluation was performed on all four benchmark datasets used in our study. The results are depicted in Tables~\ref{tab:x_step_attack_nodeinjection}(a)-(d).

% \subsubsection{Impact of Feature Perturbation Step Size ($x_{\text{step}}$) under Label Flipping Attack} \label{subsubsec:xstep_label_flipping}

We next explored the effect of feature perturbation step size under a label flipping attack. In this setting, labels of 10\%, 30\%, and 50\% of the nodes were randomly flipped to incorrect classes. 
The targeted nodes were selected based on their degrees, with the highest-degree nodes chosen for label corruption. The goal was to evaluate how the model copes with noisy label information when feature perturbation is applied with varying $x_{\text{step}}$ values. These experiments were also performed using the GCN architecture, and the feature privacy budget ($\epsilon_x$) was fixed at 1 throughout. 
As with the node injection setup, the evaluation was conducted across all four datasets. The results are depicted in Tables~\ref{tab:x_step_attack_labelflipping}(a)-(d).

\begin{table*}[t]
\centering
\caption{Impact of $x_{\text{step}}$ on Node Injection Attack Performance: Accuracy (\%) vs. $x_{\text{step}}$ for different injection rates}
\label{tab:x_step_attack_nodeinjection}

\begin{subtable}[b]{0.30\textwidth}
\centering
\caption{Cora}
\begin{tabular}{rccc}
\toprule
$x_{\text{step}}$ & 10\% & 30\% & 50\% \\
\midrule
0  & 55.60 & 58.49 & 58.09 \\
2  & 71.82 & 73.97 & 73.32 \\
4  & 75.17 & 76.94 & 75.66 \\
8  & 76.38 & 77.61 & 78.85 \\
16 & 78.39 & 79.23 & 79.00 \\
\bottomrule
\end{tabular}
\end{subtable}
\hfill
\begin{subtable}[b]{0.22\textwidth}
\centering
\caption{pubmed}
\begin{tabular}{ccc}
\toprule
10\% & 30\% & 50\% \\
\midrule
58.73 & 58.64 & 59.38 \\
75.22 & 75.19 & 76.01 \\
78.21 & 78.34 & 79.05 \\
79.97 & 80.00 & 81.02 \\
80.40 & 80.53 & 81.50 \\
\bottomrule
\end{tabular}
\end{subtable}
\hfill
\begin{subtable}[b]{0.22\textwidth}
\centering
\caption{facebook}
\begin{tabular}{ccc}
\toprule
10\% & 30\% & 50\% \\
\midrule
66.72 & 67.33 & 65.28 \\
79.88 & 81.10 & 79.12 \\
82.43 & 83.31 & 81.89 \\
83.60 & 84.67 & 83.25 \\
83.61 & 85.09 & 83.32 \\
\bottomrule
\end{tabular}
\end{subtable}
\hfill
\begin{subtable}[b]{0.22\textwidth}
\centering
\caption{lastFM}
\begin{tabular}{ccc}
\toprule
10\% & 30\% & 50\% \\
\midrule
45.44 & 49.95 & 44.23 \\
68.81 & 67.52 & 64.09 \\
72.42 & 73.96 & 72.82 \\
77.01 & 76.74 & 77.25 \\
74.35 & 75.86 & 74.90 \\
\bottomrule
\end{tabular}
\end{subtable}
\end{table*}

\begin{table*}[t]
\centering
\caption{Impact of $x_{\text{step}}$ on Label Flipping Attack Performance: Accuracy (\%) vs. $x_{\text{step}}$ for different flipping rates}
\label{tab:x_step_attack_labelflipping}

\begin{subtable}[b]{0.30\textwidth}
\centering
\caption{Cora}
\begin{tabular}{rccc}
\toprule
$x_{\text{step}}$ & 10\% & 30\% & 50\% \\
\midrule
0  & 51.24 & 41.21 & 31.86 \\
2  & 64.92 & 50.41 & 36.50 \\
4  & 67.87 & 52.38 & 38.30 \\
8  & 69.44 & 52.56 & 39.13 \\
16 & 70.19 & 53.56 & 39.69 \\
\bottomrule
\end{tabular}
\end{subtable}
\hfill
\begin{subtable}[b]{0.22\textwidth}
\centering
\caption{pubmed}
\begin{tabular}{ccc}
\toprule
10\% & 30\% & 50\% \\
\midrule
54.99 & 47.57 & 40.05 \\
69.48 & 56.62 & 44.17 \\
72.09 & 58.30 & 44.53 \\
73.57 & 59.31 & 44.97 \\
74.03 & 59.55 & 45.13 \\
\bottomrule
\end{tabular}
\end{subtable}
\hfill
\begin{subtable}[b]{0.22\textwidth}
\centering
\caption{facebook}
\begin{tabular}{ccc}
\toprule
10\% & 30\% & 50\% \\
\midrule
60.79 & 50.09 & 38.79 \\
72.59 & 58.22 & 43.23 \\
75.23 & 59.85 & 44.10 \\
76.48 & 60.93 & 44.49 \\
76.59 & 60.87 & 44.74 \\
\bottomrule
\end{tabular}
\end{subtable}
\hfill
\begin{subtable}[b]{0.22\textwidth}
\centering
\caption{lastFM}
\begin{tabular}{ccc}
\toprule
10\% & 30\% & 50\% \\
\midrule
41.02 & 32.95 & 24.61 \\
59.66 & 47.31 & 33.64 \\
65.34 & 50.02 & 36.01 \\
65.74 & 52.17 & 37.13 \\
66.20 & 52.06 & 36.64 \\
\bottomrule
\end{tabular}
\end{subtable}
\end{table*}

% \subsubsection{Impact of Label Perturbation Step Size ($y_{\text{step}}$) under Node Injection Attack} 
% \label{subsubsec:ystep_node_injection}

To further evaluate the impact of perturbation intensity on model robustness, we conducted experiments under a node injection attack scenario, this time varying the label perturbation step size ($y_{\text{step}}$). Specifically, we injected new nodes into the graph at 10\%, 30\%, and 50\% of the total number of nodes. 
The injected nodes were connected to existing high-degree nodes, and their labels were perturbed based on different values of $y_{\text{step}}$.
All the experiments were carried out using the GCN architecture. The feature privacy budget ($\epsilon_x$) was kept constant at 1, and the feature perturbation step size ($x_{\text{step}}$) was fixed for all settings. 
The evaluation was performed on all the four benchmark datasets used in our study. The results are depicted in Tables~\ref{tab:y_step_attack_nodeinjection}(a)-(d).

% \subsubsection{Impact of Label Perturbation Step Size ($y_{\text{step}}$) under Label Flipping Attack} 
% \label{subsubsec:ystep_label_flipping}

We also explored the effect of label perturbation step size ($y_{\text{step}}$) under a label flipping attack, with the highest-degree nodes targeted for corruption. The goal was to assess the model's robustness to noisy labels introduced with varying values of $y_{\text{step}}$. As with the node injection experiment, the GCN architecture was used, and the feature privacy budget ($\epsilon_x$) was fixed at 1 throughout. 
The evaluation was conducted across all the four datasets. The results are depicted in Tables~\ref{tab:y_step_attack_labelflipping}(a)-(d).
In both node injection and label flipping attacks, we observe that neither the feature perturbation step size ($x_{\text{step}}$) nor the label perturbation step size ($y_{\text{step}}$) has a significant impact on the performance of the model. This suggests that varying the intensity of feature or label perturbation alone, while keeping other factors constant, does not noticeably influence the robustness of the model under these attack scenarios.

\begin{table*}[t]
\centering
\caption{Impact of $y_{\text{step}}$ on Node Injection Attack Performance: Accuracy (\%) vs. $y_{\text{step}}$ for different injection rates}
\label{tab:y_step_attack_nodeinjection}
\begin{subtable}[b]{0.30\textwidth}
\centering
\caption{Cora}
\begin{tabular}{rccc}
\toprule
$y_{\text{step}}$ & 10\% & 30\% & 50\% \\
\midrule
0  & 72.66 & 72.11 & 72.05 \\
2  & 76.87 & 76.44 & 75.47 \\
4  & 76.79 & 78.85 & 76.28 \\
8  & 75.17 & 76.94 & 75.66 \\
16 & 69.57 & 69.75 & 70.13 \\
\bottomrule
\end{tabular}
\end{subtable}
\hfill
\begin{subtable}[b]{0.22\textwidth}
\centering
\caption{pubmed}
\begin{tabular}{ccc}
\toprule
10\% & 30\% & 50\% \\
\midrule
80.30 & 79.92 & 80.93 \\
79.59 & 79.90 & 80.23 \\
79.42 & 79.53 & 80.25 \\
78.21 & 78.34 & 79.05 \\
75.98 & 76.42 & 76.66 \\
\bottomrule
\end{tabular}
\end{subtable}
\hfill
\begin{subtable}[b]{0.22\textwidth}
\centering
\caption{facebook}
\begin{tabular}{ccc}
\toprule
10\% & 30\% & 50\% \\
\midrule
83.62 & 84.55 & 83.10 \\
87.58 & 87.95 & 86.94 \\
86.08 & 87.32 & 85.81 \\
82.43 & 83.31 & 81.89 \\
78.46 & 78.65 & 77.45 \\
\bottomrule
\end{tabular}
\end{subtable}
\hfill
\begin{subtable}[b]{0.22\textwidth}
\centering
\caption{lastFM}
\begin{tabular}{ccc}
\toprule
10\% & 30\% & 50\% \\
\midrule
72.08 & 71.43 & 66.19 \\
76.11 & 74.41 & 74.23 \\
75.11 & 74.80 & 75.74 \\
72.42 & 73.96 & 72.82 \\
70.80 & 70.30 & 70.36 \\
\bottomrule
\end{tabular}
\end{subtable}
\end{table*}
\begin{table*}[t]
\centering
\caption{Impact of $y_{\text{step}}$ on Label Flipping Attack Performance: Accuracy (\%) vs. $y_{\text{step}}$ for different flipping rates}
\label{tab:y_step_attack_labelflipping}
\begin{subtable}[b]{0.3\textwidth}
\centering
\caption{Cora}
\begin{tabular}{rccc}
\toprule
$y_{\text{step}}$ & 10\% & 30\% & 50\% \\
\midrule
0  & 62.26 & 45.41 & 29.78 \\
2  & 68.38 & 51.94 & 37.12 \\
4  & 68.91 & 52.38 & 39.23 \\
8  & 67.87 & 52.38 & 38.30 \\
16 & 61.27 & 49.32 & 36.93 \\
\bottomrule
\end{tabular}
\end{subtable}
\hfill
\begin{subtable}[b]{0.22\textwidth}
\centering
\caption{pubmed}
\begin{tabular}{ccc}
\toprule
10\% & 30\% & 50\% \\
\midrule
72.76 & 56.33 & 41.33 \\
73.06 & 58.08 & 42.92 \\
72.97 & 58.59 & 44.18 \\
72.09 & 58.30 & 44.53 \\
69.95 & 57.01 & 44.04 \\
\bottomrule
\end{tabular}
\end{subtable}
\hfill
\begin{subtable}[b]{0.22\textwidth}
\centering
\caption{facebook}
\begin{tabular}{ccc}
\toprule
10\% & 30\% & 50\% \\
\midrule
74.75 & 56.71 & 38.72 \\
79.18 & 62.10 & 44.41 \\
78.49 & 61.73 & 44.83 \\
75.23 & 59.85 & 44.10 \\
71.84 & 57.36 & 42.37 \\
\bottomrule
\end{tabular}
\end{subtable}
\hfill
\begin{subtable}[b]{0.22\textwidth}
\centering
\caption{lastFM}
\begin{tabular}{rccc}
\toprule
10\% & 30\% & 50\% \\
\midrule
60.90 & 44.69 & 27.76 \\
65.52 & 50.94 & 36.68 \\
65.19 & 51.60 & 36.28 \\
65.34 & 50.02 & 36.01 \\
59.89 & 47.56 & 34.03 \\
\bottomrule
\end{tabular}
\end{subtable}
\end{table*}

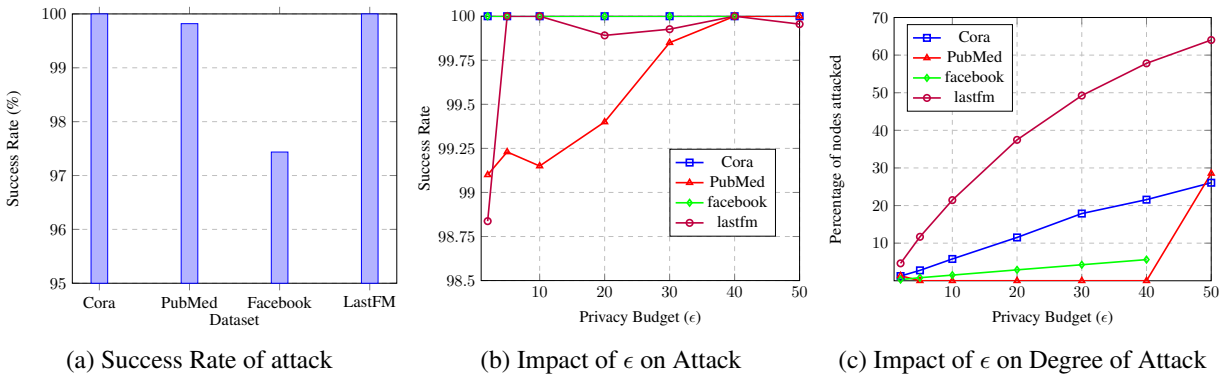
\begin{figure*}[t]
\advance\leftskip-0.00cm
 \resizebox{1\textwidth}{!}{
 \centering
\begin{subfigure}[b]{0.3\textwidth}
    \resizebox{\textwidth}{!}{
    \begin{tikzpicture}
    \begin{groupplot}[
            group style={
                group size=2 by 1,
                horizontal sep=1cm,
            },
            ]
            \nextgroupplot[
                legend columns=1,
                xlabel={Dataset},
                ylabel={Success Rate (\%)},
                ymin=95, ymax=100,
                ybar,
                bar width=10pt,
                ytick={95,96,97,98,99,100},
                xticklabels={Cora, PubMed, Facebook, LastFM},
                xtick=data,
                legend pos=south east,
                ymajorgrids=true,
                % xmajorgrids=true,
                grid style=dashed,
                legend style={at={(axis cs:0.5,96)},anchor=south west},
            ]
            \addplot coordinates {(0,100) (1, 99.819) (2, 97.4359) (3, 100)};
            % \addlegendentry{Non-Adaptive}

        \end{groupplot}
    \end{tikzpicture}}
     \caption{Success Rate of attack} 
    \label{fig:gcn_label_flipping_attack_results}
 \end{subfigure}
 \hfill
 \begin{subfigure}[b]{0.3\textwidth}
    \resizebox{\textwidth}{!}{
    \begin{tikzpicture}
    \begin{axis}[
        legend columns=1,
        xlabel={Privacy Budget ($\epsilon$)},
        ylabel={Success Rate},
        xmin=1, xmax=50,
        ymin=98.5, ymax=100,
        ytick={98.5,98.75,99,99.25,99.5,99.75,100},
        xtick={0,10,20,30,40,50},
        legend pos=south east,
        ymajorgrids=true,
        xmajorgrids=true,
        grid style=dashed,
        legend style={at={(axis cs:30,98.75)},anchor=south west},
    ]

    \addplot[
        color=blue,
        mark=square,
        line width=1.0pt
        ]
        coordinates {
        (2, 100) (5, 100) (10,100) (20, 100) (30, 100) (40, 100) (50, 100)
        
        };
        \addlegendentry{Cora}

    \addplot[
        color=red,
        mark=triangle,
        line width=1.0pt
        ]
        coordinates {
        (2, 99.1) (5, 99.23) (10,99.15) (20, 99.4) (30, 99.85) (40, 100) (50, 100)
        
        };
        \addlegendentry{PubMed}
    \addplot[
        color=green,
        mark=diamond,
        line width=1.0pt
        ]
        coordinates {
        (2, 100) (5, 100) (10,100) (20, 100) (30, 100) (40, 100) 
        
        };
        \addlegendentry{facebook}
    \addplot[
        color=purple,
        mark=o,
        line width=1.0pt
        ]
        coordinates {
        (2, 98.8372) (5, 100) (10, 100) (20, 99.8912) (30, 99.9264) (40, 100) (50,99.9552)
        };
        \addlegendentry{lastfm}

    % \addplot[
    %     color=green,
    %     mark=diamond,
    %     line width=1.0pt
    %     ]
    %     coordinates { %LATEST
    %     (2, 100) (5, 100) (10,100) (20, 100) (30, 100) (40, 100) (50, 100)
        
    %     };
    %     \addlegendentry{Facebook}

    \end{axis}
    \end{tikzpicture}}
     \caption{Impact of $\epsilon$ on Attack} 
    \label{fig:epsilon_poisoning_attack_results}
 \end{subfigure}
 \hfill
 \begin{subfigure}[b]{0.3\textwidth}
    \resizebox{\textwidth}{!}{
    \begin{tikzpicture}
    % \begin{groupplot}[
    %         group style={
    %             group size=2 by 1,
    %             horizontal sep=1cm,
    %         },
    %         ]
    %         \nextgroupplot[
    %             legend columns=1,
    %             xlabel={Dataset},
    %             ylabel={Percentage of susceptible nodes attacked},
    %             ymin=0, ymax=100,
    %             ybar,
    %             bar width=10pt,
    %             ytick={0,10,20,30,40,50,60,70,80,90,100},
    %             xticklabels={Cora, PubMed, Facebook, LastFM},
    %             xtick=data,
    %             % xtick={1,2,3,4,5,6,7,8,9,10},
    %             legend pos=south east,
    %             ymajorgrids=true,
    %             % xmajorgrids=true,
    %             grid style=dashed,
    %             legend style={at={(axis cs:0.5,20)},anchor=south west},
    %         ]
    %         \addplot coordinates {(0, 0.27) (1, 0.111) (2, 0.3) (3, 0.5)};
    %         % \addlegendentry{Non-Adaptive}

    %     \end{groupplot}
    \begin{axis}[
        legend columns=1,
        xlabel={Privacy Budget ($\epsilon$)},
        ylabel={Percentage of nodes attacked},
        xmin=1, xmax=50,
        ymin=0, ymax=70,
        ytick={10,20,30,40,50,60,70},
        xtick={0,10,20,30,40,50},
        legend pos=south east,
        ymajorgrids=true,
        xmajorgrids=true,
        grid style=dashed,
        legend style={at={(axis cs:3,45)},anchor=south west},
    ]

    \addplot[
        color=blue,
        mark=square,
        line width=1.0pt
        ]
        coordinates { %LATEST
        (2, 1.255) (5, 2.767) (10,5.797) (20, 11.5214) (30, 17.87) (40, 21.567) (50, 26.07)
        
        };
        \addlegendentry{Cora}

    \addplot[
        color=red,
        mark=triangle,
        line width=1.0pt
        ]
        coordinates { %LATEST
        (2, 1.334) (5, 0) (10,0) (20, 0) (30, 0) (40, 0) (50, 28.554)
        
        };
        \addlegendentry{PubMed}
    \addplot[
        color=green,
        mark=diamond,
        line width=1.0pt
        ]
        coordinates { %LATEST
        (2, 0.3115) (5, 0.80996) (10,1.5086) (20, 2.9016) (30, 4.259) (40, 5.6252) 
        };
        \addlegendentry{facebook}

    \addplot[
        color=purple,
        mark=o,
        line width=1.0pt
        ]
        coordinates { %LATEST
        (2, 4.6308) (5, 11.6899) (10,21.4457) (20, 37.4699) (30, 49.2446) (40, 57.8144)(50, 64.0124) 
        };
        \addlegendentry{lastfm}

    % \addplot[
    %     color=green,
    %     mark=diamond,
    %     line width=1.0pt
    %     ]
    %     coordinates { %LATEST
    %     (2, 100) (5, 100) (10,100) (20, 100) (30, 100) (40, 100) (50, 100)
        
    %     };
    %     \addlegendentry{Facebook}

    \end{axis}
    \end{tikzpicture}}
 
    \caption{Impact of $\epsilon$ on Degree of Attack} 
    \label{fig:sage_label_flipping_attack_results}
 \end{subfigure}
 \hfill
 % \caption{Experimental Results}
 % \label{fig:ExpResults}
 }
 \caption{Success Rate of the Data Poisoning Attack and the impact of privacy budget ($\epsilon$)}
 \label{fig:poisoning_attack_results}
\end{figure*} 
\subsection{Data Poisoning Attack Results}
\label{subsec:data_poisoning_results}
From the results in Tables \ref{tab:x_step_attack_nodeinjection}-\ref{tab:y_step_attack_labelflipping}, it can be concluded that the effectiveness of the attack strategies remains distinct. Label flipping continues to show a strong impact, with noticeable accuracy degradation as the percentage of flipped labels increases. In contrast, node injection still does not significantly degrade performance, indicating its limited effectiveness under the evaluated conditions.

\subsection{LPGNN Inference Attack Results}
\label{subsec:GNN_inference_attack_results}

We also conducted the inference attacks on LPGNN as outlined in Section \ref{subsec:gnn_inference_attack}. For these experiments, two metrics, namely cosine similarity and mean feature difference, were employed to gain insight into how well the attack fairs against the LDP guarantee of the multi-bit mechanism. 
Our attack primarily aimed at nodes with higher degree, and the results are shown in Table \ref{tab:inferenceattackresults}. Both cosine similarity (C. Similarity) values and the mean feature difference (Mean F.D.) values point towards a very poor attack performance. A value closer to $0$ for cosine similarity implies that the predicted feature vector and original feature vector are neither very similar, nor very different from each other. It is also relevant to note that the mean feature difference value is considerably greater than $1$ when the domain for the original feature values itself is $(0,1)$ in these datasets. This implies that the difference being reported between feature values predicted and original feature values is more than the size of the feature value domain itself. This is a puzzling discovery, but we can explain the reason for this occurrence.

The primary reason for the failure of inference attacks is the vastness of the output domain after applying the multi-bit mechanisms. Upon thorough analysis, we found that although the original feature value domain is $[0,1]$, after the multi-bit mechanisms are applied, those same values are hidden inside the domain $[-178.5,179.5]$, which is a vast domain difference. This diversity in the final output domain of the LDP mechanism causes a stronger and more accurate privacy protection of the values, leading to unsuccessful inference attacks. It is important to note that this attack would work perfectly under the premise that the feature values are all very similar in a given close neighborhood, but this is not always true. This also led to a poor attack performance. 
Therefore, we conclude that an inference attack based on collective aggregation of unbiased noisy values would not work well in a locally private GNN setting.

% \vspace{0.1in}

\begin{table}[t]
\begin{center}
\caption{Inference Attack Results}
\label{tab:inferenceattackresults}
 \begin{tabular}{ |p{2.0cm}||p{0.9cm}|p{1.1cm}|p{1.3cm}|p{1cm}|  }
 \hline
 \multicolumn{5}{|c|}{LPGNN Inference Attack Results} \\
 \hline
 Metric & Cora & PubMed & Facebook & LastFM\\
 \hline
 C. Similarity   & 0.005 & 0.001 & 0.001 & 0.001\\
 Mean F. D. & 21.973 & 7.924 & 66.924 & 116.666\\
 \hline
\end{tabular}
\end{center}
\end{table}

Finally, we tested our data poisoning attack on the four datasets and the attack was able to achieve as high as $100\%$ accuracy in its inferences and the results are shown in Figures \ref{fig:poisoning_attack_results}(a)-(c). We followed the inference attack after the poisoning, as explained in Section \ref{subsec:data_poisoning_results}. Our attack always concluded that whenever the response from multi-bit encoder was $x^*=1$ for a feature, the original feature value must have been $x_i=1$, as Cora dataset has the feature range $[0,1]$ with only two possible feature values - $0$ and $1$. 

Our experimental findings support the theoretical basis for this attack as introduced in Section \ref{proof:poisoning_attack}, making it a very strong attack against locally private graph neural networks.
With the Cora dataset, we could conduct absolute inference attacks due to its binary-valued features. 
% We achieved $100\%$ success-rate with our attack on the Cora dataset. 
With the PubMed dataset, our attack was able to achieve $99.82\%$ success rate. The attack also performed well against other datasets - $97.44\%$ against Facebook and $100\%$ against LastFM. 
It is most important to note that with this successful inference attack, we are able to breach the local differential privacy guarantee offered by the multi-bit mechanisms and therefore compromise the privacy of the nodes in the graph. Therefore, our data poisoning attack against LPGNN is a strong success.

% \textbf{Defense against this attack}: 
% As seen from the results above, our data poisoning attack proved to be extremely successful in breaking the LDP guarantees offered by the multi-bit mechanism in LPGNN, highlighting the need for a reliable defense against it. 
% We now discuss a possible defense mechanism that would allow the nodes to identify their feature values have been poisoned and thus not fall prey to the attacker. Note that after the poison is added, the feature value goes below the agreed upon domain $(\alpha,\beta)$, specifically if the value was $x_i=\alpha$. This in turn rigs the Bernoulli trial that happens in Line 4 of the multi bit encoder algorithm (Algorithm \ref{algorithm:mbm}). We can mitigate the risk of this attack if the nodes are programmed to validate their feature values and check if they fall within $(\alpha,\beta)$. Hence the nodes would be able to identify the poisoned values that could potentially leak information, and act accordingly.

\section{Related Work}
\label{sec:related}

The vulnerability of Graph Neural Networks to adversarial perturbations has attracted significant attention in recent years. However work on the vulnerability of Locally Private Graph Neural Nets is limited and yet to be explored in depth. Early work  \cite{Zugner_2018} demonstrated that slight modifications to a graph’s structure or node features could dramatically alter GNN predictions.
Dai et al. \cite{adversarials_on_gnn} investigate adversarial attacks by proposing a meta-learning based approach for the task of node classification. Their framework exploits meta-gradients to solve the bilevel optimization problem underlying the challenging class of poisoning adversarial attacks, even transfering to unsupervised models. 

Zhang et al. \cite{surveyOnPrivacyAttacksInGNNs} do a comprehensive review of the privacy considerations related to graph data and models, discussing attacks such as model extraction attack, graph structure reconstruction, attribute inference attack and membership inference attack. A recent survey \cite{gnn_surveyOnLinKbwPrivacySecurity} consolidated most of the latest work on security and privacy for GNNs, summarizing both adversarial attacks and adversarial defenses. 
Further, Hsieh et al. \cite{netfense} provide an adversarial defense - a graph perturbation-based approach named NetFense to keep graph data unnoticeability, maintain the prediction confidence of targeted label classification, and reduce the prediction confidence of private label classification. Zuegner et al. \cite{adversarials_on_gnn} propose an algorithm called NETTACK, exploiting incremental computations and drastically dropping the accuracy of GNNs by minimal perturbations. 
In contrast, Ma et al. \cite{attackAsInfluenceMaxProblem} address perturbation attacks as an influence maximization problem on the graph, drawing a formal analysis. In another work \cite{practicaladversarialattacksGNN}, the same authors study black-box attacks on graph neural networks under practical constraints. In a competing approach, Zhang et al. \cite{defendAttackviaTensorEnhance} employ defense against adversarial attacks by using tensor approximation, systematically aggregating and compressing diverse predefined robustness features of adversarial graphs into a low-rank representation.

% However, all the work reviewed above have only focus on the vulnerabilities of traditional graph neural networks, establishing the fact that they possess vulnerabilities that attackers can misuse to compromise performance or privacy. 
Some of the more recent research presented many alternative solutions, combining various defensive mechanisms to forge stronger, private versions of graph neural networks, or their training processes. For example, Chien et al. \cite{chien2023differentiallyprivatedecoupledgraph} introduced DP-GNN, a framework embedding differential privacy into the training of GNNs. Their work also introduced a framework termed Graph Differential Privacy (GDP), specifically tailored to graph learning. 
Mueller et al. \cite{DP-GNN_forGraphClassification} introduced a framework for differentially private graph-level classification, relying on the differentially private stochastic gradient descent (DP-SGD) method, similar to another recent work \cite{trainingdifferentiallyprivategraphwithrandomwalksampling}. The latter utilize disjoint sub-graphs to train, and propose three random-walk based methods for generating such sub-graphs. 
However, all of these build upon the traditional model of Differential Privacy, which is known to be weaker than LDP.
The work of Olatunji et al. \cite{releaseGNNWithDP} approach this problem differently by proposing PrivGNN, a framework to release GNN models trained on sensitive data in a privacy-preserving manner. However, it still cannot prevent the privacy risks if the server conducting the GNN training is compromised and not trustworthy. 
Two of the existing approaches (\cite{lpgnn} and \cite{ldpgraph}) explore integrating LDP into the learning frameworks for GNNs. Joshi and Mishra \cite{LocallyStructurePrivateGNN} further explore the requirement of preserving the privacy of graph structure, in addition to the node features and labels. Another work \cite{secGNN} implement a system supporting privacy-preserving GNN training and inference in the cloud, using knowledge of light-weight cryptography and ML techniques. There are also several novel applications of privacy preserving GNNs \cite{surveyOnPrivacyAttacksInGNNs}\cite{progap}.

% Collectively, all the above work underscore the emerging intersection between adversarial robustness and privacy in graph-based deep learning. While traditional adversarial attack strategies expose fundamental weaknesses of GNNs, the integration of local privacy mechanisms introduces novel considerations that call for further research into robust and privacy-preserving design. Our work is the first to study attacks on locally private graph neural networks, to the best of our knowledge.

\section{Conclusion}
\label{sec:concl}
We have presented the first comprehensive study of adversarial attacks on LPGNNs, a privacy-preserving GNN framework built upon LDP. We conducted extensive experiments on four benchmark datasets—Cora, PubMed, Facebook, and LastFM—across multiple GNN architectures (GCN, GAT, and GraphSAGE). We evaluated the impact of four core adversarial strategies: Node Injection, Label Flipping, Data Inference, and Data Poisoning. Some direction towards defense against data poisoning attacks is given in the Appendix.

We believe our work paves the way for the design of GNN architectures that are not only grounded in strong theoretical privacy guarantees but also demonstrate practical resilience against a wide array of adversarial threats. We hope that this study will serve as a foundation for further exploration at the intersection of privacy, security and graph-based deep learning. Application specific attacks and their countermeasures will constitute an important aspect of such explorations. Looking ahead, we would like to explore additional dimensions of adversarial robustness in LPGNNs. Other datasets and GNN architectures can also be included in the experiments. Modeling adversarial attacks and defenses in LPGNN as a game will be an interesting new direction of research. 
% Specifically, we plan to simulate surrogate GNN attacks, where an adversary trains a separate GNN model to mimic the behavior of the target LPGNN and uses this approximation to generate transferable attacks. This will allow us to evaluate black-box attack scenarios, which are more realistic in privacy-preserving settings.

% \clearpage

% \bibliographystyle{ACM-Reference-Format}
\bibliographystyle{splncs04}
\bibliography{main}

@String{Computing = "Computing" }

@String{Computer = "{IEEE} Computer" }

@String{Springer = "Springer-Verlag" }

@article{kipf_gnn_gcn,
  author       = {Thomas N. Kipf and
                  Max Welling},
  title        = {Semi-Supervised Classification with Graph Convolutional Networks},
  journal      = {CoRR},
  volume       = {abs/1609.02907},
  year         = {2016},
  url          = {http://arxiv.org/abs/1609.02907},
  eprinttype    = {arXiv},
  eprint       = {1609.02907},
  bibsource    = {dblp computer science bibliography, https://dblp.org}
}

@misc{veličković2018_gans,
      title={Graph Attention Networks}, 
      author={Petar Veličković and Guillem Cucurull and Arantxa Casanova and Adriana Romero and Pietro Liò and Yoshua Bengio},
      year={2018},
      eprint={1710.10903},
      archivePrefix={arXiv},
      primaryClass={stat.ML}
}

@article{adversarials_on_gnn,
author = {Z\"{u}gner, Daniel and Borchert, Oliver and Akbarnejad, Amir and G\"{u}nnemann, Stephan},
title = {Adversarial Attacks on Graph Neural Networks: Perturbations and their Patterns},
year = {2020},
issue_date = {October 2020},
volume = {14},
number = {5},
issn = {1556-4681},
journal = {ACM Trans. Knowl. Discov. Data},
month = {jun},
articleno = {57},
numpages = {31}
}

@misc{graph_injection_attack,
      title={Understanding and Improving Graph Injection Attack by Promoting Unnoticeability}, 
      author={Yongqiang Chen and Han Yang and Yonggang Zhang and Kaili Ma and Tongliang Liu and Bo Han and James Cheng},
      year={2022},
      eprint={2202.08057},
      archivePrefix={arXiv},
      primaryClass={cs.LG}
}

@article{Algorithmic_foundations_of_DP,
author = {Dwork, Cynthia and Roth, Aaron},
title = {The Algorithmic Foundations of Differential Privacy},
year = {2014},
issue_date = {August 2014},
volume = {9},
number = {3–4},
issn = {1551-305X},
journal = {Found. Trends Theor. Comput. Sci.},
month = {aug},
pages = {211–407},
numpages = {197}
}

@article{lpgnn,
  author       = {Sina Sajadmanesh and
                  Daniel Gatica{-}Perez},
  title        = {Locally Private Graph Neural Networks},
  journal      = {CoRR},
  volume       = {abs/2006.05535},
  year         = {2021},
  url          = {https://arxiv.org/abs/2006.05535},
  eprinttype    = {arXiv},
  eprint       = {2006.05535},
  bibsource    = {dblp computer science bibliography, https://dblp.org}
}

@inproceedings{Zugner_2018,
   title={Adversarial Attacks on Neural Networks for Graph Data},
   booktitle={Proceedings of the 24th ACM SIGKDD International Conference on Knowledge Discovery \& Data Mining},
   author={Zügner, Daniel and Akbarnejad, Amir and Günnemann, Stephan},
   year={2018},
   pages = {2847–2856},
   month=jul }

@ARTICLE{surveyOnPrivacyAttacksInGNNs,

  author={Zhang, Yi and Zhao, Yuying and Li, Zhaoqing and Cheng, Xueqi and Wang, Yu and Kotevska, Olivera and Yu, Philip S. and Derr, Tyler},

  journal={IEEE Transactions on Knowledge and Data Engineering}, 

  title={A Survey on Privacy in Graph Neural Networks: Attacks, Preservation, and Applications}, 

  year={2024},

  volume={36},

  number={12},

  pages={7497-7515}}

@article{gnn_surveyOnLinKbwPrivacySecurity,
author = {Guan, Faqian and Zhu, Tianqing and Zhou, Wanlei and Choo, Kim-Kwang},
year = {2024},
month = {02},
pages = {},
title = {Graph neural networks: a survey on the links between privacy and security},
volume = {57},
journal = {Artificial Intelligence Review}
}

@ARTICLE{DP-GNN_forGraphClassification,

  author={Mueller, Tamara T. and Paetzold, Johannes C. and Prabhakar, Chinmay and Usynin, Dmitrii and Rueckert, Daniel and Kaissis, Georgios},

  journal={IEEE Transactions on Pattern Analysis and Machine Intelligence}, 

  title={Differentially Private Graph Neural Networks for Whole-Graph Classification}, 

  year={2023},

  volume={45},

  number={6},

  pages={7308-7318}}

@ARTICLE{secGNN,

  author={Wang, Songlei and Zheng, Yifeng and Jia, Xiaohua},

  journal={IEEE Transactions on Services Computing}, 

  title={SecGNN: Privacy-Preserving Graph Neural Network Training and Inference as a Cloud Service}, 

  year={2023},

  volume={16},

  number={4},

  pages={2923-2938}}

@ARTICLE{modelInversaionAgainstGNN,

  author={Zhang, Zaixi and Liu, Qi and Huang, Zhenya and Wang, Hao and Lee, Chee-Kong and Chen, Enhong},

  journal={IEEE Transactions on Knowledge and Data Engineering}, 

  title={Model Inversion Attacks Against Graph Neural Networks}, 

  year={2023},

  volume={35},

  number={9},

  pages={8729-8741}}

@misc{releaseGNNWithDP,
      title={Releasing Graph Neural Networks with Differential Privacy Guarantees}, 
      author={Iyiola E. Olatunji and Thorben Funke and Megha Khosla},
      year={2023},
      eprint={2109.08907},
      archivePrefix={arXiv},
      primaryClass={cs.LG},
      url={https://arxiv.org/abs/2109.08907}, 
}

@article{kasiviswanathan2011can,
  title={What can we learn privately?},
  author={Kasiviswanathan, Shiva Prasad and Lee, Homin K and Nissim, Kobbi and Raskhodnikova, Sofya and Smith, Adam},
  journal={SIAM Journal on Computing},
  volume={40},
  number={3},
  pages={793--826},
  year={2011}
}

@article{rappor,
  author       = {{\'{U}}lfar Erlingsson and
                  Aleksandra Korolova and
                  Vasyl Pihur},
  title        = {{RAPPOR:} Randomized Aggregatable Privacy-Preserving Ordinal Response},
  journal      = {CoRR},
  volume       = {abs/1407.6981},
  year         = {2014},
  url          = {http://arxiv.org/abs/1407.6981},
  eprinttype    = {arXiv},
  eprint       = {1407.6981},
  bibsource    = {dblp computer science bibliography, https://dblp.org}
}

@inproceedings{dwork2006calibrating,
  title={Calibrating noise to sensitivity in private data analysis},
  author={Dwork, Cynthia and McSherry, Frank and Nissim, Kobbi and Smith, Adam},
  booktitle={Theory of cryptography conference},
  pages={265--284},
  year={2006},
  organization={Springer}
}

@InProceedings{Privacy_via_dist_noise_gen,
author="Dwork, Cynthia
and Kenthapadi, Krishnaram
and McSherry, Frank
and Mironov, Ilya
and Naor, Moni",
editor="Vaudenay, Serge",
title="Our Data, Ourselves: Privacy Via Distributed Noise Generation",
booktitle="Advances in Cryptology - EUROCRYPT 2006",
year="2006",
pages="486--503",
isbn="978-3-540-34547-3"
}

@ARTICLE{gnnPaper,

  author={Scarselli, Franco and Gori, Marco and Tsoi, Ah Chung and Hagenbuchner, Markus and Monfardini, Gabriele},

  journal={IEEE Transactions on Neural Networks}, 

  title={The Graph Neural Network Model}, 

  year={2009},

  volume={20},

  number={1},

  pages={61-80}}

@ARTICLE{netfense,

  author={Hsieh, I-Chung and Li, Cheng-Te},

  journal={IEEE Transactions on Knowledge and Data Engineering}, 

  title={NetFense: Adversarial Defenses Against Privacy Attacks on Neural Networks for Graph Data}, 

  year={2023},

  volume={35},

  number={1},

  pages={796-809}}

@misc{chien2023differentiallyprivatedecoupledgraph,
      title={Differentially Private Decoupled Graph Convolutions for Multigranular Topology Protection}, 
      author={Eli Chien and Wei-Ning Chen and Chao Pan and Pan Li and Ayfer Özgür and Olgica Milenkovic},
      year={2023},
      eprint={2307.06422},
      archivePrefix={arXiv},
      primaryClass={cs.LG},
      url={https://arxiv.org/abs/2307.06422}, 
}

@misc{ldpgraph,
      title={Local Differential Privacy in Graph Neural Networks: a Reconstruction Approach}, 
      author={Karuna Bhaila and Wen Huang and Yongkai Wu and Xintao Wu},
      year={2024},
      eprint={2309.08569},
      archivePrefix={arXiv},
      primaryClass={cs.LG},
      url={https://arxiv.org/abs/2309.08569}, 
}

@article{LocallyStructurePrivateGNN,
author = {Joshi, Rucha Bhalchandra and Mishra, Subhankar},
title = {Locally and Structurally Private Graph Neural Networks},
year = {2024},
issue_date = {March 2024},
number = {1},
journal = {Digital Threats},
month = mar,
articleno = {10},
numpages = {23}
}

@misc{trainingdifferentiallyprivategraphwithrandomwalksampling,
      title={Training Differentially Private Graph Neural Networks with Random Walk Sampling}, 
      author={Morgane Ayle and Jan Schuchardt and Lukas Gosch and Daniel Zügner and Stephan Günnemann},
      year={2023},
      eprint={2301.00738},
      archivePrefix={arXiv},
      primaryClass={cs.LG},
      url={https://arxiv.org/abs/2301.00738}, 
}

@misc{attackAsInfluenceMaxProblem,
      title={Adversarial Attack on Graph Neural Networks as An Influence Maximization Problem}, 
      author={Jiaqi Ma and Junwei Deng and Qiaozhu Mei},
      year={2021},
      eprint={2106.10785},
      archivePrefix={arXiv},
      primaryClass={cs.LG},
      url={https://arxiv.org/abs/2106.10785}, 
}

@misc{practicaladversarialattacksGNN,
      title={Towards More Practical Adversarial Attacks on Graph Neural Networks}, 
      author={Jiaqi Ma and Shuangrui Ding and Qiaozhu Mei},
      year={2021},
      eprint={2006.05057},
      archivePrefix={arXiv},
      primaryClass={cs.LG},
      url={https://arxiv.org/abs/2006.05057}, 
}

@inproceedings{effectiveNodeInjectionAttacks,
author = {Zou, Xu and Zheng, Qinkai and Dong, Yuxiao and Guan, Xinyu and Kharlamov, Evgeny and Lu, Jialiang and Tang, Jie},
title = {TDGIA: Effective Injection Attacks on Graph Neural Networks},
year = {2021},
booktitle = {Proceedings of the 27th ACM SIGKDD Conference on Knowledge Discovery \& Data Mining},
pages = {2461–2471},
numpages = {11},
location = {Virtual Event, Singapore},
series = {KDD '21}
}

@inproceedings{singleNodeInjectionAttacks,
author = {Tao, Shuchang and Cao, Qi and Shen, Huawei and Huang, Junjie and Wu, Yunfan and Cheng, Xueqi},
title = {Single Node Injection Attack against Graph Neural Networks},
year = {2021},
booktitle = {Proceedings of the 30th ACM International Conference on Information \& Knowledge Management},
pages = {1794–1803},
numpages = {10},
location = {Virtual Event, Queensland, Australia},
series = {CIKM '21}
}

@inproceedings{advAttackOnGNNviaNodeInjections,
author = {Sun, Yiwei and Wang, Suhang and Tang, Xianfeng and Hsieh, Tsung-Yu and Honavar, Vasant},
title = {Adversarial Attacks on Graph Neural Networks via Node Injections: A Hierarchical Reinforcement Learning Approach},
year = {2020},
booktitle = {Proceedings of The Web Conference 2020},
pages = {673–683},
numpages = {11},
location = {Taipei, Taiwan},
series = {WWW '20}
}

@ARTICLE{GANI_NodeInjections,

  author={Fang, Junyuan and Wen, Haixian and Wu, Jiajing and Xuan, Qi and Zheng, Zibin and Tse, Chi K.},

  journal={IEEE Transactions on Computational Social Systems}, 

  title={GANI: Global Attacks on Graph Neural Networks via Imperceptible Node Injections}, 

  year={2024},

  volume={11},

  number={4},

  pages={5374-5387}}

@article{defendAttackviaTensorEnhance,
title = {Defending adversarial attacks in Graph Neural Networks via tensor enhancement},
journal = {Pattern Recognition},
volume = {158},
pages = {110954},
year = {2025},
author = {Jianfu Zhang and Yan Hong and Dawei Cheng and Liqing Zhang and Qibin Zhao}
}

@article{classificationOptNodeInjectionAttack,
title = {Classification optimization node injection attack on graph neural networks},
journal = {Knowledge-Based Systems},
volume = {301},
pages = {112323},
year = {2024},
author = {Mingda Ma and Hui Xia and Xin Li and Rui Zhang and Shuo Xu}
}

@INPROCEEDINGS{advLabelFlippingAttacknDefence,

  author={Zhang, Mengmei and Hu, Linmei and Shi, Chuan and Wang, Xiao},

  booktitle={2020 IEEE International Conference on Data Mining (ICDM)}, 

  title={Adversarial Label-Flipping Attack and Defense for Graph Neural Networks}, 

  year={2020},

  volume={},

  number={},

  pages={791-800}}

@INPROCEEDINGS{multiLabelFlippingAdvAttackOnGNN,

  author={Li, Jianfeng and Li, Haoran and He, Jing and Dou, Tianchen},

  booktitle={2023 International Seminar on Computer Science and Engineering Technology (SCSET)}, 

  title={Lapa: Multi-Label-Flipping Adversarial Attacks on Graph Neural Networks}, 

  year={2023},

  volume={},

  number={},

  pages={29-32},

  doi={10.1109/SCSET58950.2023.00016}}

@ARTICLE{ieee_tp,
  author={De Chaudhury, Saptarshi and Morreddigari, Likhith Reddy and Varun, Matta and Sengupta, Tirthankar and Chakraborty, Sandip and Sural, Shamik and Vaidya, Jaideep and Atluri, Vijayalakshmi},
  journal={IEEE Transactions on Privacy}, 
  title={Blockchain Based Secure Federated Learning With Local Differential Privacy and Incentivization}, 
  year={2024},
  volume={1},
  number={},
  pages={31-44}}

@article{gnn_Explain,
author = {Li, Xuyan and Wang, Jie and Yan, Zheng},
title = {Can Graph Neural Networks be Adequately Explained? A Survey},
year = {2025},
volume = {57},
number = {5},
journal = {ACM Comput. Surv.},
month = jan,
articleno = {131},
numpages = {36}
}

@inproceedings{progap,
author = {Sajadmanesh, Sina and Gatica-Perez, Daniel},
title = {ProGAP: Progressive Graph Neural Networks with Differential Privacy Guarantees},
year = {2024},
booktitle = {Proceedings of the 17th ACM International Conference on Web Search and Data Mining},
pages = {596–605},
numpages = {10},
location = {Merida, Mexico},
series = {WSDM '24}
}

\appendix
\section{Appendix}
\label{sec:appendix}
This Appendix contains some additional material related to the
work presented in this paper.
\subsection{Defense against Data Poisoning attack}
\label{subsec:appendix_defense}
As seen from the results presented in Section \ref{sec:experiments}, our data poisoning attack proved to be extremely successful in breaking the LDP guarantees offered by the multi-bit mechanism in LPGNN, highlighting the need for a reliable defense against it. 
We now discuss a possible defense mechanism that would allow the nodes to identify their feature values have been poisoned and thus not fall prey to the attacker. Note that after the poison is added, the feature value goes below the agreed upon domain $(\alpha,\beta)$, specifically if the value was $x_i=\alpha$. This in turn rigs the Bernoulli trial that happens in Line 4 of the multi bit encoder algorithm (Algorithm \ref{algorithm:mbm}). We can mitigate the risk of this attack if the nodes are programmed to validate their feature values and check if they fall within $(\alpha,\beta)$. Hence the nodes would be able to identify the poisoned values that could potentially leak information, and act accordingly.








\end{document}